\documentclass{article}

\usepackage[utf8]{inputenc}
\usepackage[T1]{fontenc}
\usepackage{hyperref}
\usepackage{url}
\usepackage{booktabs}
\usepackage{amsfonts}
\usepackage{nicefrac}
\usepackage{microtype}
\usepackage{xcolor}
\usepackage{amsmath}
\usepackage{amssymb}
\usepackage{mathtools}
\usepackage{wrapfig}
\usepackage{amsthm}
\usepackage{multirow}
\usepackage{graphicx}
\usepackage{natbib}
\usepackage[margin=1in]{geometry}
\usepackage{algorithm}
\usepackage{algorithmic}
\usepackage{float}
\usepackage{bm}
\usepackage{authblk}

\theoremstyle{definition}
\newtheorem{definition}{Definition}[section]
\newtheorem{proposition}{Proposition}[section]

\title{Beyond Entropy: Learning from Token-Level Distributional Deviations for LLM Reasoning}

\author[1]{Xuanzhi Feng}
\author[1]{Zhengyang Li$^{*}$}
\author[1]{Zeyu Liu$^{*}$}
\author[1]{Haoxi Li}
\author[2]{Yuming Jiang}
\author[2]{Bing Guo}
\author[3]{Jingcai Guo}
\author[1]{Jie Zhang$^{\dagger}$}
\author[1]{Song Guo$^{\dagger}$}

\affil[1]{The Hong Kong University of Science and Technology, Hong Kong, China}
\affil[2]{Sichuan University, Chengdu, China}
\affil[3]{The Hong Kong Polytechnic University, Hong Kong, China}

\date{}

\begin{document}

\maketitle

\begin{center}
\small
\textsuperscript{*}Equal contribution. \quad
\textsuperscript{$\dagger$}Corresponding authors: Jie Zhang (jzhang@ust.hk), Song Guo (songguo@ust.hk).
\end{center}

\begin{abstract}
Reinforcement Learning with Verifiable Rewards (RLVR) has significantly advanced Large Language Model (LLM) reasoning; however, it faces a fundamental optimization instability: uniform token updates precipitate entropy collapse, leading to premature convergence to suboptimal strategies, whereas excessive Shannon Entropy maximization can cause entropy explosion, driving blind exploration toward incoherent reasoning chains. To resolve this dichotomy, we introduce the Independent Combinatorial Tokens (ICT) framework, which shifts the optimization focus from scalar uncertainty to the distributional properties of token logits. By leveraging the Jensen-Shannon (JS) divergence between token logits distributions, ICT identifies tokens with distinctive distributional patterns as critical branching points for guiding effective exploration in LLM reasoning. Our theoretical analysis, grounded in both Shannon and second-order R\'{e}nyi entropy, proves that selectively updating on these tokens regulates policy concentration: it reduces the overall distribution uncertainty measured by Shannon entropy, while controlling probability concentration captured by second-order R\'{e}nyi entropy. This dual effect prevents over-concentrated token generation from weakening exploration and effectively stabilizes the training landscape. Empirical results demonstrate that updating only the top 10\% of unique tokens on Qwen2.5 (0.5B/1.5B/7B) models yields an average pass@4 improvement of 4.58\%, with a maximum gain of 14.9\%, over GRPO, 20-Entropy, and STAPO baselines across seven benchmarks spanning math, commonsense, and Olympiad-level problems.
\end{abstract}

\begin{figure}[t]
  \centering
  \includegraphics[width=0.7\columnwidth]{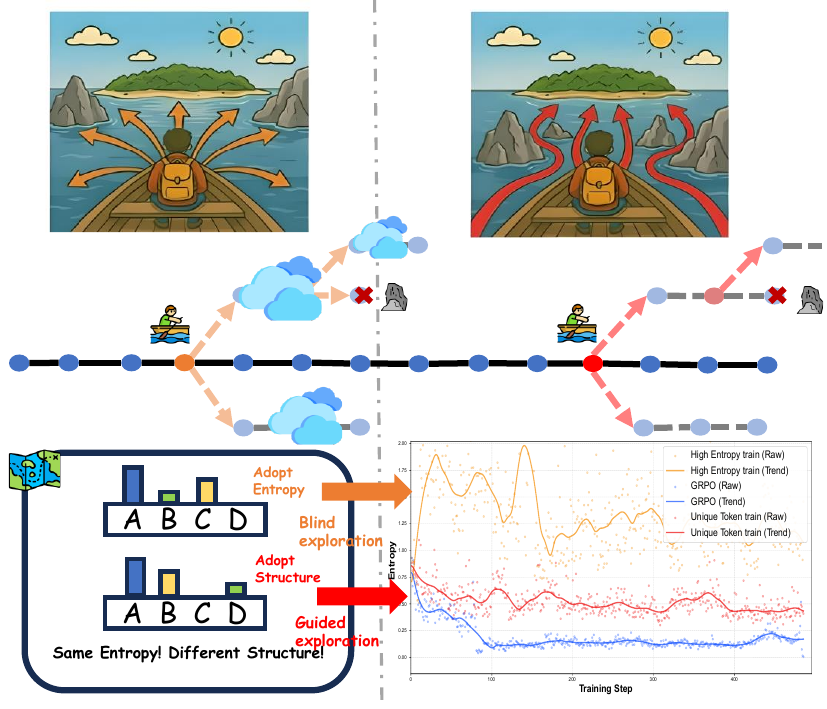}
  \caption{(a) Blind Exploration (Shannon Entropy): Different probability distributions can yield the same Shannon entropy, making the signal ambiguous. This leads to blind exploration, often resulting in dead ends and persistent uncertainty (orange curve).
    (b) Guided Exploration (Distributional Information): The ICT framework identifies critical branching points among promising paths via distributional properties. This enables guided exploration, achieving stable convergence (red curve) while avoiding Shannon-entropy instability and entropy collapse from uniform token updates (blue curve).
  }
  \label{Fig1}
\end{figure}

\section{Introduction}
\label{sec:Introduction}
Large Language Models (LLMs) have demonstrated remarkable capabilities in mathematical and programmatic reasoning with Reinforcement Learning with Verifiable Rewards (RLVR) \cite{liu2024deepseek,guo2025deepseek}. Despite these gains, prevailing implementations often apply training signals uniformly across all tokens \cite{chen2025acereason,wang2025reinforcement}, which results in premature convergence and entropy collapse, thereby diminishing the model's exploration capacity \cite{shao2024deepseekmath}. To address this challenge, researchers have sought to enhance exploration by identifying and rewarding critical tokens that serve as key points in reasoning trajectories \cite{lin2024critical,sun2025ktae,ruan2025enhancing}. Current methods \cite{wang2025beyond,wang2025entropy,cao2025efficient} mainly rely on Shannon Entropy to select high-entropy tokens as logical branching points, while recent studies \cite{jiang2025rethinking,cao2025efficient} show that Shannon Entropy merely reflects uncertainty without indicating future exploration directions; consequently, empirical findings \cite{wang2025beyond,wang2025entropy} suggest that focusing exclusively on high-entropy tokens often leads to overly dispersed and blind exploration.

Although existing methods attempt to mitigate entropy collapse and explosion, most of them rely on direct monitoring \cite{hao2025rethinking,zhang2025rediscovering} or heuristic constraints \cite{potraghloo2025top} to artificially regulate the entropy change, failing to address the root cause. From an information-theoretic perspective, Shannon Entropy serves merely as a proxy for uncertainty and is not the fundamental driver of effective exploration \cite{wang2025arbitrary,jin2025revisiting}. Critically, probability distributions with identical Shannon Entropy can be vastly different, leading to divergent inference trajectories \cite{yang2025not}, as illustrated in Figure \ref{Fig1}. Therefore, relying solely on a scalar entropy value to assess token importance is insufficient.

To address these limitations, we investigate the underlying mechanisms of RLVR from the perspective of the distributional properties of token logits. Motivated by a fundamental principle of information theory -- the self-information of an event is inversely proportional to its probability \cite{shannon1948mathematical,yeung1991new} -- we posit that a token's uniqueness correlates with the informational importance it encodes. Extending this principle to autoregressive language modeling, we define unique tokens as those whose logit distribution exhibits the greatest deviation from the sequence-level average distribution. By quantifying this deviation via the Jensen-Shannon (JS) divergence between individual token logit distributions, we identify tokens that serve as critical branching points, providing essential directional guidance for effective exploration.

Based on this insight, we propose the Independent Combinatorial Tokens (ICT) framework.
ICT guides pivotal exploration by selectively updating the gradients of these unique tokens, preventing both entropy collapse and explosion and thereby preserving stable exploration capacity. 
Theoretically, we prove that these distributionally unique tokens act as critical branching points that encode valuable exploration directions; selectively updating them enables stable and effective exploration while avoiding Shannon Entropy-induced instabilities.

Quantitative experiments on Qwen2.5 models (0.5B to 7B) demonstrate that retaining gradients for only the top 10\% of distributionally unique tokens achieves performance comparable to, or higher than, full gradient updates. This strategy's effectiveness scales with model size and represents, to our knowledge, the first approach that achieves implicit entropy control and enhanced exploration without directly optimizing entropy metrics.

We further validate ICT via sparse policy gradient updates: only the top 10\% of distributionally unique tokens are updated, while the remaining 90\% are masked. Evaluated on three base models (0.5B, 1.5B, and 7B), this sparse strategy matches or exceeds the performance of full-parameter training, despite using only 10\% of the gradients. These results indicate that a small subset of distributionally unique tokens, acting as key decision nodes, contributes the vast majority of the performance gains in RLVR.

In summary, we highlight the pivotal role of unique tokens in both reasoning and exploration. 
This work establishes a token-centric foundation for the next generation of RLVR algorithms. Our main contributions are as follows:
\begin{itemize}
\item \textbf{Distributional Analysis of Logits:} We propose identifying unique tokens via the Jensen-Shannon (JS) divergence between logits distributions, shifting the focus from Shannon Entropy to the distributional properties of logits. This demonstrates that critical reasoning branches are governed by distributional properties rather than scalar uncertainty alone.
\item \textbf{ICT Framework:} We introduce the \textbf{Independent Combinatorial Tokens (ICT)} framework, which selectively updates high-information unique tokens. ICT provides directional guidance for exploration while preventing both entropy collapse and explosion.
\item \textbf{Performance:} We demonstrate that updating only the top 10\% of distributionally unique tokens suffices to match or exceed the performance of full-parameter training. Experiments on Qwen2.5 models series across seven benchmarks verify that ICT scales effectively. Compared with strong baselines including GRPO, 20-Entropy, and STAPO, our method yields an average pass@4 improvement of 4.37\%, 4.98\% and 4.38\%, with maximum gains reaching up to 14.9\%.
\end{itemize}

\section{Preliminaries}
\subsection{Measures of Uncertainty}

\textbf{Shannon Entropy.}
Shannon entropy quantifies the uncertainty of the policy distribution $\pi_\theta(\cdot \mid o_{<t})$ at each generation step corresponding to token $o_t$. It is formally defined as
\[
\mathcal{H}_1(\pi_\theta \mid o_{<t}) = -\sum_{a \in \mathcal{V}} \pi_\theta(a \mid o_{<t}) \log \pi_\theta(a \mid o_{<t}),
\]
where $\mathcal{V}$ denotes the vocabulary and $o_{<t}$ represents the sequence of preceding tokens. In standard reinforcement learning, maximizing Shannon entropy promotes exploration by encouraging a more uniform distribution over the action (token) space in the absence of strong external rewards.

We argue that Shannon entropy is suboptimal for LLMs, whose vocabularies are extremely large and exhibit heavy-tailed distributions. Instead, we adopt the second-order R\'{e}nyi entropy $\mathcal{H}_2$, which is derived from the collision probability and more effectively captures the concentration of probability mass in the policy distribution.

\begin{definition}[Collision Probability and R\'{e}nyi Entropy]
The collision probability of the policy at step $t$ is
\[
C_t(\theta) = \sum_{a \in \mathcal{V}} \pi_\theta(a \mid o_{<t})^2.
\]
The second-order R\'{e}nyi entropy is then
\[
\mathcal{H}_2(\pi_\theta \mid o_{<t}) = -\log C_t(\theta).
\]
\end{definition}

A higher value of $\mathcal{H}_2$ indicates greater effective diversity across tokens, while a lower value reflects a more concentrated and deterministic distribution. 
Unlike Shannon entropy, $\mathcal{H}_2$ is less sensitive to low-probability noise in the long tail of the vocabulary because of its quadratic dependence on the probabilities ($\sum \pi^2$). By suppressing the influence of aleatoric noise, $\mathcal{H}_2$ provides a more robust approximation of the effective support size of the decision space, enabling the reliable identification of distributionally significant tokens that are critical for the exploration--exploitation trade-off in LLM reasoning.

\subsection{Jensen--Shannon divergence}
Traditional self-information $I(o_t) := -\log \pi_\theta(o_t \mid o_{<t})$ fails to distinguish the functional roles of different tokens in a reasoning trajectory. We therefore adopt a distributional perspective and quantify information as the deviation of the policy distribution $\boldsymbol{p}_t = \pi_\theta(\cdot \mid o_{<t})$ from the uniform distribution $\boldsymbol{u} = \mathbb{1}/|\mathcal{V}|$ via Kullback--Leibler divergence:
\[
D_{\mathrm{KL}}(\boldsymbol{p}_t \parallel \boldsymbol{u}) = \sum_{i=1}^{|\mathcal{V}|} p_{t,i} \log(|\mathcal{V}| \cdot p_{t,i}).
\]
For improved numerical stability and symmetry, we further employ the Jensen--Shannon (JS) divergence between the policy distribution $\boldsymbol{p}$ and a reference distribution $\boldsymbol{q}$:
\[
\mathrm{JS}(\boldsymbol{p} \parallel \boldsymbol{q}) := \frac{1}{2} D_{\mathrm{KL}}(\boldsymbol{p} \parallel \boldsymbol{m}) + \frac{1}{2} D_{\mathrm{KL}}(\boldsymbol{q} \parallel \boldsymbol{m}),
\]
where $\boldsymbol{m} = \frac{1}{2}(\boldsymbol{p} + \boldsymbol{q})$. This metric is bounded in $[0, \log 2]$, ensuring stable evaluation of distributional shifts across tokens.

\subsection{Group Relative Policy Optimization}
We adopt Group Relative Policy Optimization (GRPO)~\cite{shao2024deepseekmath} as the backbone RLVR algorithm. Given a prompt $q$, the current policy $\pi_\theta$ samples $G$ output sequences $\{o_i\}_{i=1}^G$, where $o_i = (o_{i,1}, o_{i,2}, \dots, o_{i,T_i})$. A verifiable reward $r_i$ (e.g., answer correctness) is computed for each sequence. The advantage $A_i$ is then calculated by normalizing the rewards within the group: $A_i = \frac{r_i - \mathrm{mean}(\{r_j\})}{\mathrm{std}(\{r_j\})}$.

The standard GRPO loss function is:
\vspace{-0.2em}
\begin{equation}
\small
\mathcal{J}_{\text{GRPO}}(\theta) = \mathbb{E}_{q,\{o_i\}\sim\pi_{\text{old}}}\left[ \frac{1}{G} \sum_{i=1}^G \frac{1}{T_i} \sum_{t=1}^{T_i} \Psi_{i,t}(\theta) \right],
\end{equation}
\vspace{-0.15em}
where the per-token objective $\Psi_{i,t}(\theta)$ consists of a clipped probability ratio and a KL penalty:
\vspace{-0.15em}
\begin{equation}
\small
\Psi_{i,t}(\theta) = \min\!\left( \rho_{i,t} A_i,\; \operatorname{clip}(\rho_{i,t}, 1-\epsilon, 1+\epsilon) A_i \right) - \beta' \, D_{\text{KL}}\bigl(\pi_\theta(\cdot\mid q,o_{i,<t}) \;\|\; \pi_{\text{ref}}(\cdot\mid q,o_{i,<t})\bigr).
\end{equation}
\vspace{-0.15em}
Here $\rho_{i,t} = \frac{\pi_\theta(o_{i,t}\mid q,o_{i,<t})}{\pi_{\text{old}}(o_{i,t}\mid q,o_{i,<t})}$, $\pi_{\text{ref}}$ is a reference policy (typically the initial checkpoint), and $\beta'$ controls the KL penalty strength. This formulation enables stable policy updates while avoiding over-optimization of the verifiable reward.

In the following section, we show that uniformly applying this objective to all tokens leads to entropy instability, and we propose a sparse variant that focuses updates on unique tokens.

\section{Unique Tokens Drive Effective Exploration}

\begin{figure}[t]
  \centering
  \includegraphics[width=\textwidth]{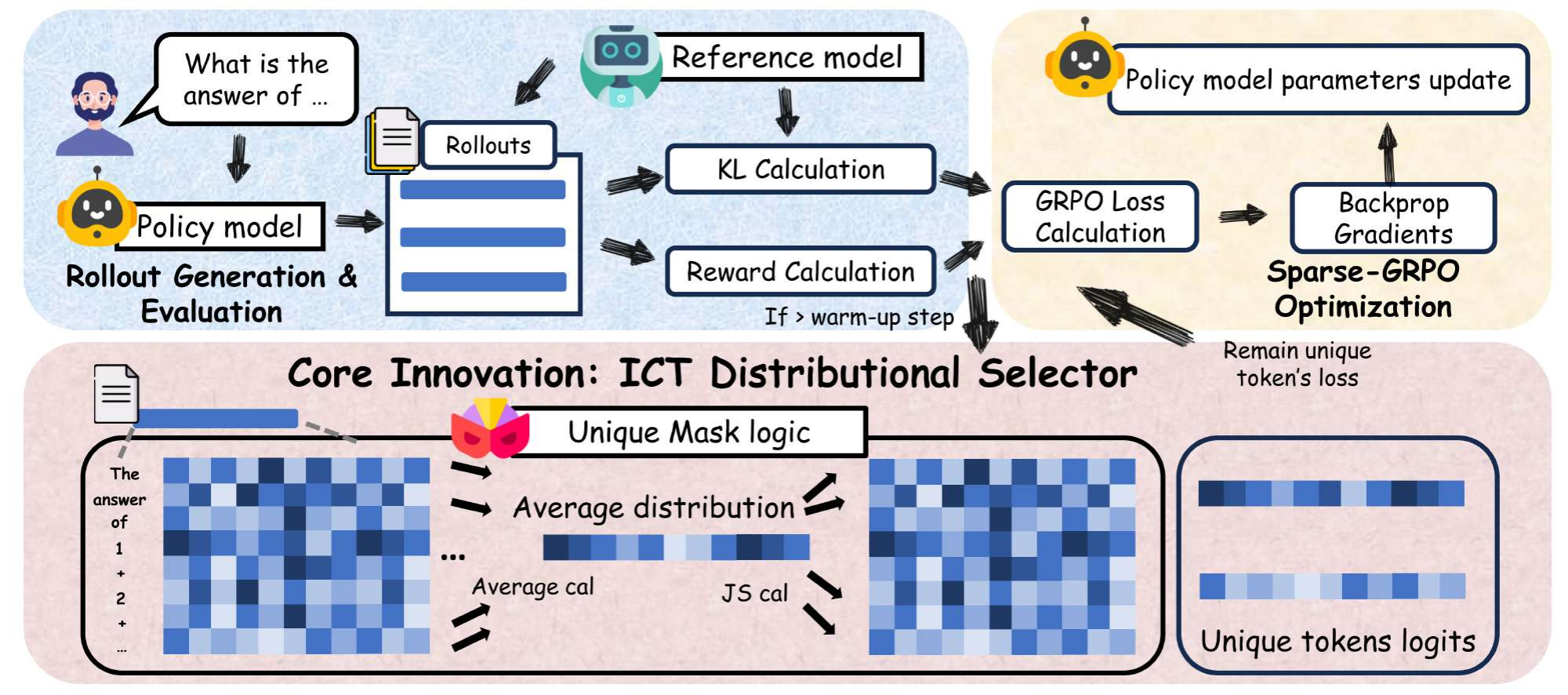}
  \caption{ICT-based Sparse RLVR Framework}
  \label{Fig2}
\end{figure}

As highlighted in Section \ref{sec:Introduction}, uniform token-level updates often induce entropy collapse~\cite{shao2024deepseekmath}, whereas training exclusively on high-entropy tokens risks entropy explosion~\cite{wang2025beyond}. These opposing phenomena, which we validate in Figure \ref{Fig1}, substantially impair the model's exploration capacity. To address this issue, we introduce the Independent Combinatorial Tokens (ICT) framework (Figure \ref{Fig2}), which selectively updates unique tokens to maintain a balance between exploration and convergence.

In this section, we provide theoretical grounding for the ICT framework. We address the instability of RLVR by formalizing the entropy dynamics under two opposing optimization regimes. Rather than applying uniform token updates, we categorize updates according to their impact on the distributional properties of the policy. To do so, we first characterize the conditions that lead to the extremes of entropy collapse and explosion.

We define the \emph{strategy purity} $\beta(\pi)$ as the collision probability of the current policy, serving as a dynamic threshold for stability. Based on this, we introduce the following regimes:

\textbf{Regime of Entropy Collapse.} Let $\mathcal{R}_H := \{o_t \mid \pi(o_t) > \beta(\pi)\}$ denote the set of high-confidence tokens, where the token probability exceeds the strategy purity threshold. Continuous optimization within $\mathcal{R}_H$ yields a negative entropy gradient ($\Delta \mathcal{H}_2 < 0$). Unrestricted updates in this regime force the policy $\pi_\theta$ to degenerate into a deterministic distribution centered on local optima, thereby eliminating exploration.

\textbf{Regime of Entropy Explosion.} Let $\mathcal{R}_L := \{o_t \mid \pi(o_t) < \beta(\pi)\}$ denote the set of low-confidence tokens (the long tail). Optimization focused on $\mathcal{R}_L$ yields a positive entropy gradient ($\Delta \mathcal{H}_2 > 0$). Excessive updates in this regime cause the policy distribution to flatten towards uniformity, leading to catastrophic forgetting of the prompt context and divergent reasoning paths.

Standard full-token training induces opposing dynamics between these two regimes. We formally state that the objective of the ICT framework is to resolve this conflict by selectively updating \emph{unique tokens}---those situated at the critical branching points between $\mathcal{R}_H$ and $\mathcal{R}_L$---thereby maintaining the entropy equilibrium and enabling stable exploration.

\subsection{Theoretical Analysis: Entropy Dynamics and Unique Tokens Update}

\label{app:proof_homogeneity}
Entropy dynamics typically refer to changes in Shannon entropy~\cite{wang2025entropy,wang2025beyond}. However, as introduced in the Preliminaries, Shannon entropy alone is insufficient to fully characterize the model's exploration capacity. The second-order R\'{e}nyi entropy $\mathcal{H}_2$ serves as a more robust indicator of model uncertainty and further reflects the model's exploration behavior. Therefore, we first analyze the relationship between changes in second-order R\'{e}nyi entropy and unique token updates, followed by the relationship between Shannon entropy and unique token updates.

To analyze how unique token updates influence the system's exploration capacity, we examine the gradient dynamics of the second-order R\'{e}nyi entropy $\mathcal{H}_2$. The dynamics of $\mathcal{H}_2$ are governed by a re-weighted policy distribution that prioritizes high-probability regions, which we refer to as the squared policy.

\textbf{Squared Policy Distribution.} For a given policy $\pi_\theta(\cdot|s)$ and its associated collision probability $C(\theta)$, the squared policy $p_2(\cdot|s)$ is defined as the probability distribution proportional to the square of the original policy probabilities:
\vspace{-0.2em}
\begin{equation}
\small
    p_2(a|s) := \frac{\pi_\theta(a|s)^2}{C(\theta)} = \frac{\pi_\theta(a|s)^2}{\sum_{a' \in \mathcal{A}} \pi_\theta(a'|s)^2}.
\end{equation}
\vspace{-0.2em}
The squared policy $p_2$ serves as a sharpening mechanism. Due to the quadratic dependence, $p_2$ significantly amplifies the probability mass of dominant modes while suppressing the long tail of low-probability tokens. This ensures that the optimization of $\mathcal{H}_2$ is driven by the most competitive candidates rather than aleatoric noise.

\textbf{Gradient of $\mathcal{H}_2$.} Leveraging the squared policy, the gradient of the second-order R\'{e}nyi entropy with respect to parameters $\theta$ is given by the expected score function under $p_2$:
\vspace{-0.2em}
\begin{equation}
    \small
    \nabla_\theta \mathcal{H}_2(\pi_\theta|s) = -2 \mathbb{E}_{a \sim p_2(\cdot|s)} \left[ \nabla_\theta \log \pi_\theta(a|s) \right].
\end{equation}
\vspace{-0.2em}
This establishes that the sensitivity of $\mathcal{H}_2$ is governed by $p_2(a|s) \propto \pi_\theta(a|s)^2$. To explicitly quantify the impact of updating a specific token on the model's exploration capacity, we isolate the normalization constant of this distribution. This term, which represents the aggregate concentration of probability mass, serves as a critical baseline. We formally define this metric as \emph{strategy purity}.

\textbf{Strategy Purity.} The strategy purity of the current policy, denoted $\beta(\pi)$, is its collision probability:
\vspace{-0.2em}
\begin{equation}
    \small
    \beta(\pi) = \sum_{a \in \mathcal{A}} \pi(a)^2.
\end{equation}
\vspace{-0.2em}
\textbf{Entropy Dynamics under Sparse Updates.} With this metric established, we can now derive the precise effect of a sparse parameter update on the system's uncertainty. Assume an update is applied to a specific token $a^*$ (tokens with high JS divergence from the group average are selected; higher-order interactions are negligible under small learning rates, hence we use a single token \(a^*\)~\cite{meng2026sparse} to represent) such that its logit increases by a small amount $\Delta\theta_{s, a^*} > 0$. The first-order approximation of the change in second-order entropy $\Delta \mathcal{H}_2$ is given by:
\vspace{-0.2em}
\begin{equation}
    \small
    \Delta \mathcal{H}_2 \approx -2 \Delta\theta_{s, a^*} \pi(a^*) \left( \frac{\pi(a^*)}{\beta(\pi)} - 1 \right).
\end{equation}
\vspace{-0.2em}

Consequently, the sign of $\Delta \mathcal{H}_2$ is determined solely by the relationship between the token probability $\pi(a^*)$ and the strategy purity $\beta(\pi)$. This reveals a critical bifurcation in the training process. (Caveat on First-Order Approximation: The linear approximation in Eq.~(6) assumes an infinitesimally small step size $\Delta\theta$, i.e., proximity of policy updates. In practice, aggressive learning rates or large advantage spikes can cause substantial parameter jumps, making higher-order Taylor residuals non-negligible. Furthermore, AdamW's momentum introduces historical inertia into $\Delta\theta$. But in practice the learning rate is sufficiently small, ensuring the first-order approximation remains mathematically valid and dominates over momentum-induced deviations.)

\textbf{Entropy Bifurcation.} The update dynamics exhibit two distinct regimes based on the token's probability relative to the distribution's purity:
\label{Entropy Bifurcation}
\begin{enumerate}
    \item \textbf{Regime H (Entropy Collapse):} If $\pi(a^*) > \beta(\pi)$, then $\Delta \mathcal{H}_2 < 0$. Updating high-probability tokens decreases entropy.
    \item \textbf{Regime L (Entropy Expansion):} If $\pi(a^*) < \beta(\pi)$, then $\Delta \mathcal{H}_2 > 0$. Updating low-probability tokens increases entropy.
\end{enumerate}

In practical implementations, the L2 regularization imposed by optimizers such as AdamW~\cite{zhou2024towards} constrains the magnitude of logits, encouraging a symmetric distribution of token probabilities. This constraint helps maintain a balance between the populations of high-confidence tokens in $\mathcal{R}_H$ and low-confidence tokens in $\mathcal{R}_L$ during training.

This analysis elucidates the instability inherent in full-token training: the model is subjected to conflicting entropy gradients from high- and low-confidence tokens. By identifying and updating unique tokens---which occupy the critical branching points between these two regimes---we effectively mitigate the opposing risks of collapse and explosion. This selective update strategy leverages the natural equilibrium provided by standard optimizers while providing precise directional guidance for stable exploration.

\textbf{Homogeneity of Two Entropy Measures.} Let $\pi(a)$ denote the probability of token $a$ under the policy $\pi_\theta(\cdot|s)$. For any dominant token $a^*$ satisfying the high-confidence condition $\pi(a^*) > e^{-1}$, the gradients of the Shannon entropy $\mathcal{H}_1$ and the second-order R\'{e}nyi entropy $\mathcal{H}_2$ with respect to $\pi(a^*)$ point in the same direction:
\vspace{-0.2em}
\begin{equation}
    \small
    \mathrm{sgn}\left(\frac{\partial \mathcal{H}_1}{\partial \pi(a^*)}\right) = \mathrm{sgn}\left(\frac{\partial \mathcal{H}_2}{\partial \pi(a^*)}\right) = -1.
\end{equation}
\vspace{-0.2em}
This implies that increasing the confidence in dominant tokens simultaneously minimizes both entropy measures.

We analyze the sensitivity of both entropy measures with respect to the probability mass of a dominant token. By computing the partial derivatives, we show that under the high-confidence condition ($\pi(a) > e^{-1}$), the gradients of both measures are negative. The detailed derivation is provided in Appendix~\ref{app:proof_homogeneity}.

\subsection{Sparse Policy Gradient by ICT Distributional Selector}
Transitioning from the theoretical analysis of entropy dynamics to a practical algorithm, we formulate a tractable policy gradient objective. While the theory indicates that updating unique tokens is key to balancing exploration and convergence, standard methods indiscriminately optimize every generated token.

We introduce the Independent Combinatorial Tokens (ICT) framework, which refines Group Relative Policy Optimization (GRPO) by incorporating an ICT distributional selector. This selector identifies unique tokens at critical branching points and constructs a sparse mask to filter gradients, thereby ensuring that the model focuses solely on high-information learning signals.

\textbf{Deriving the Sparse Objective.} We build upon the standard GRPO formulation. Consider a group of $G$ outputs $\{o_i\}_{i=1}^G$ sampled from the old policy $\pi_{\text{old}}$ given a prompt $q$. Let $\Psi_{i,t}(\theta)$ denote the standard per-token objective function at step $t$ for output $i$, which encapsulates the advantage-weighted clipped probability ratio and the KL penalty:
\vspace{-0.2em}
\begin{equation}
\small
\Psi_{i,t}(\theta) \triangleq \mathcal{L}^{\text{CLIP}}_{i,t}(\theta) - \beta D_{\mathrm{KL}}(\pi_\theta \parallel \pi_{\text{ref}})_t.
\end{equation}
\vspace{-0.2em}
In conventional GRPO, the learning objective averages $\Psi_{i,t}$ over all tokens in the sequence. However, as highlighted above, such dense supervision introduces conflicting entropy gradients from trivial tokens. To resolve this, we modulate the objective with the sparse mask generated by the ICT distributional selector.

\textbf{Independent Combinatorial Token Distributional Selector.} We employ the Jensen-Shannon (JS) divergence to quantify each token's uniqueness relative to the group-average distribution. Let $L_{i,t}$ be the logit distribution of the $i$-th token. We first compute the group's average distribution $P_{\text{avg}}(\cdot|t) = \frac{1}{G} \sum_{j=1}^G \mathrm{softmax}(L_{j,t})$. The uniqueness score $u_{i,t}$ is defined as:
\vspace{-0.2em}
\begin{equation}
\small
u_{i,t} = D_{\mathrm{JS}} \left( \mathrm{softmax}(L_{i,t}) \parallel P_{\text{avg}}(\cdot|t) \right).
\end{equation}
\vspace{-0.2em}
The ICT distributional selector then constructs a binary sparse mask $M_{i,t} \in \{0,1\}$ that retains only the tokens in the top $k$-percentile of uniqueness scores:
\vspace{-0.2em}
\begin{equation}
\small
M_{i,t} = \mathbb{I}\left[ u_{i,t} \ge \text{Percentile}(\{u_{i,\tau}\}_{\tau=1}^T, k) \right].
\end{equation}
\vspace{-0.2em}
\textbf{The Sparse-GRPO Estimator.} Substituting the sparse mask into the gradient formulation yields the \emph{Sparse-GRPO} objective. Unlike the standard GRPO objective, which averages over the full sequence length $T$, our formulation normalizes updates with respect to the support of the active mask:
\vspace{-0.2em}
\begin{equation}
\small
\mathcal{J}_{\text{S-GRPO}}(\theta) = \mathbb{E}_{q, \{o_i\}} \left[ \frac{1}{G} \sum_{i=1}^G \frac{1}{\sum_{t} M_{i,t}} \sum_{t=1}^{T_i} M_{i,t} \cdot \Psi_{i,t}(\theta) \right].
\end{equation}
\vspace{-0.2em}
This formulation ensures that optimization resources are strictly allocated to the unique tokens that drive effective exploration.

\subsection{Algorithm Implementation}

\begin{minipage}[t]{0.4\textwidth}
    \vspace{0pt}
    \hrule height 0.8pt
    \vspace{3pt}
    \footnotesize \textbf{Algorithm 1} ICT framework
    \vspace{3pt}
    \hrule height 0.4pt
    \begin{algorithmic}[1]
        \footnotesize
        \STATE {\bfseries Input:} $\mathcal{D}, G, \Theta, \alpha, k$
        \STATE {\bfseries Params:} Policy $\theta, \pi_{\text{ref}}$
        \FOR{step $s = 1, \dots, S$}
            \STATE $q \sim \mathcal{D}$; $\{o_i\}_{i=1}^G \sim \pi_{\theta}(\cdot|q)$
            \STATE Compute $r_i$ and $A_i$
            \FOR{$i \in \{1, \dots, G\}$}
                \STATE $\Psi_{i,t}(\theta) \leftarrow$ (PPO-Clip + KL)
                \IF{$s < \Theta$}
                    \STATE $M_{i,t} \leftarrow 1$
                \ELSE
                    \STATE $P_{\text{avg}} \leftarrow \frac{1}{G} \sum \mathrm{softmax}(L_{j,t})$
                    \STATE $u_{i,t} \leftarrow D_{\mathrm{JS}}(\mathrm{softmax} \parallel P_{\text{avg}})$
                    \STATE $M_{i,t} \leftarrow \mathbb{I}[u_{i,t} \ge \mathrm{Perc}]$
                \ENDIF
            \ENDFOR
            \STATE $\mathcal{L} \leftarrow - \frac{1}{G} \sum \frac{1}{\sum M} \sum M \Psi$
            \STATE $\theta \leftarrow \theta - \alpha \nabla_\theta \mathcal{L}$
        \ENDFOR
    \end{algorithmic}
    \hrule height 0.9pt
\end{minipage}
\hfill
\begin{minipage}[t]{0.60\textwidth}
    \vspace{0pt}
    \normalsize
    The resulting algorithm, \textbf{ICT}, is detailed in Algorithm 1. It incorporates a warm-up mechanism to stabilize the initial policy before enabling the sparse selector. 
    \vspace{3pt}
    In practice, the Jensen-Shannon divergence computation for the ICT distributional selector leverages parallel batch processing on the logit vectors, incurring negligible computational overhead compared to the substantial savings achieved during the sparse backward pass. More implementation details can be found in Appendix~\ref{sec:detailed_algo}.
\end{minipage}

\section{Experiment}
\subsection{Experimental Setup}
We build our training pipeline upon VeRL~\cite{sheng2025hybridflow} and closely follow the GRPO training recipe~\cite{shao2024deepseekmath}, using the mean across 5 independent random seeds. The only difference is that our method applies sparse updates exclusively to unique tokens identified by the ICT distributional selector, while keeping all other components---the loss function, optimizer, and so on---identical to the GRPO baseline. We also include 20-Entropy~\cite{wang2025beyond} and STAPO~\cite{liu2026stapo} as baselines. We train separate models on the GSM8K and MATH training sets and evaluate them on GSM8K~\cite{lightman2023lets}, Math500~\cite{lightman2023lets}, MMLU-Stem~\cite{hendrycks2020measuring}, GPQA~\cite{rein2023gpqa}, AIME23~\cite{zan2026cama}, AIME24~\cite{li2024numinamath}, and AIME25~\cite{opencompass_aime2025}. More details can be found in Appendix~\ref{Experiment Setting details}.

\subsection{Main Results}
As shown in Table \ref{tab:main_results}, we systematically evaluate the performance of the ICT method on the Qwen2.5 series models and conduct a comprehensive comparison against multiple baselines. Experimental results demonstrate that \textbf{ICT consistently achieves the highest Avg P@1, P@4, and Total scores across all model scales and seven benchmarks, delivering stable and substantial gains}. Concretely, ICT outperforms the other three baselines by 3.38\%, 4.31\%, and 3.94\% points on the 0.5B, 1.5B, and 7B models, respectively, and surpasses the Qwen2.5 base models by 6.28\%, 5.12\%, and 4.16\% average points. These findings confirm the effectiveness and robustness of the ICT model.

Although ICT is trained on mathematical datasets (Math500 and GSM8K), its \textbf{performance improvements generalize successfully to other domains} (GPQA). In contrast, alternative training paradigms suffer from generalization trade-offs that degrade overall average performance. For example, the Qwen2.5-1.5B model trained with 20-entropy gets strong average results on math benchmarks (Pass@1 = 19.81\%, Pass@4 = 26.04\%), surpassing the GRPO-trained counterpart (18.95\% and 25.66\%). However, when non-mathematical datasets are taken into account, the 20-Entropy variant exhibits a noticeable decline in average performance.

As we have emphasized in Section \ref{sec:Introduction}, \textbf{ICT can enhance the exploration of models, which emerges clearly from the differential improvements in Pass@1 versus Pass@4}. Across all three model scales, the average gain in P@4 substantially exceeds that in P@1 (0.5B: +3.37\% vs.\ +2.37\%; 1.5B: +4.98\% vs.\ +3.64\%; 7B: +4.38\% vs.\ +3.50\%). Pass@4 quantifies the probability of obtaining at least one correct solution across four samples; these larger gains indicate that ICT produces a richer and more diverse set of correct reasoning trajectories rather than redundant paths.

\begin{table}[H]
\centering
\caption{Comparison of performance across different model sizes and benchmarks.}
\label{tab:main_results}
\small
\setlength{\tabcolsep}{4pt}
\renewcommand{\arraystretch}{1.1}
\resizebox{\textwidth}{!}{%
\begin{tabular}{l*{17}{c}}
\toprule
\multirow{2}{*}{\textbf{Method}} 
& \multicolumn{2}{c}{\textbf{Math500 (\%)}} 
& \multicolumn{2}{c}{\textbf{GSM8K (\%)}} 
& \multicolumn{2}{c}{\textbf{MMLU-s (\%)}} 
& \multicolumn{2}{c}{\textbf{GPQA (\%)}} 
& \multicolumn{2}{c}{\textbf{AIME23 (\%)}} 
& \multicolumn{2}{c}{\textbf{AIME24 (\%)}} 
& \multicolumn{2}{c}{\textbf{AIME25 (\%)}}
& \multicolumn{3}{c}{\textbf{Avg \%}} \\
\cmidrule(lr){2-3} \cmidrule(lr){4-5} \cmidrule(lr){6-7} \cmidrule(lr){8-9} \cmidrule(lr){10-11} \cmidrule(lr){12-13} \cmidrule(lr){14-15} \cmidrule(lr){16-18}
& P@1 & P@4 & P@1 & P@4 & P@1 & P@4 & P@1 & P@4 & P@1 & P@4 & P@1 & P@4 & P@1 & P@4 & P@1 & P@4 & Total \\
\midrule
\multicolumn{18}{c}{\textit{\textbf{Qwen2.5-0.5B}}} \\
\midrule
Base      & 11.05 & 21.20 & 13.88 & 38.62 & 11.22 & 30.98 & 4.56 & 12.49 & 0.86 & 3.45 & 0.00 & 0.00 & 0.00 & 0.00 & 5.65  & 14.48 & 10.07 \\
GRPO      & 11.70 & 23.20 & \textbf{42.87} & 63.15 & 12.52 & 32.27 & 5.11 & 8.44  & 0.86 & 3.45 & 0.56 & 2.25 & 0.00 & 0.00 & 10.52 & 18.97 & 14.75 \\
20-Entropy& 11.35 & 22.40 & 30.38 & 50.50 & 11.51 & 31.10 & 3.98 & 10.75 & 0.86 & 3.45 & 0.00 & 0.00 & 0.00 & 0.00 & 8.26  & 16.37 & 12.32 \\
STAPO     & 10.23 & 21.50 & 40.88 & 64.50 & 12.34 & 32.20 & 2.78 & 8.37  & 0.00 & 0.00 & 0.57 & 2.27 & 0.00 & 0.00 & 9.54  & 18.29 & 13.92 \\
ICT (Ours)& \textbf{12.15} & \textbf{24.80} & 40.27 & \textbf{65.40} & \textbf{16.85} & \textbf{40.62} & \textbf{5.95} & \textbf{15.69} & 0.86 & 3.45 & 0.57 & 2.27 & 0.00 & 0.00 & \textbf{10.95} & \textbf{21.75} & \textbf{16.35} \\
 & & & & & & & & & & & & & & & {\scriptsize (+2.37)} & {\scriptsize (+4.37)} & {\scriptsize (+3.38)} \\
\midrule
\multicolumn{18}{c}{\textit{\textbf{Qwen2.5-1.5B}}} \\
\midrule
Base      & 17.25 & 31.40 & 43.10 & 77.29 & 33.95 & 64.80 & 14.02 & 28.70 & 3.45 & 6.90  & 1.42 & 6.82  & 1.67 & 6.67  & 16.41 & 31.80 & 24.11 \\
GRPO      & 17.90 & 33.80 & 61.51 & 81.71 & 29.70 & 61.40 & 13.84 & 28.91 & 4.11 & 10.21 & 0.28 & 4.56  & 0.21 & 2.47  & 18.22 & 31.87 & 25.05 \\
20-Entropy& 17.95 & 31.60 & 59.50 & 82.12 & 36.12 & 65.17 & 11.77 & 23.56 & 4.31 & 10.34 & 0.57 & 2.27  & 0.43 & 2.11  & 18.66 & 31.02 & 24.84 \\
STAPO     & 18.05 & 32.80 & 62.31 & 80.15 & 35.24 & 59.24 & 16.65 & 29.34 & 4.45 & 11.21 & 1.71 & 4.55  & 0.83 & 3.33  & 19.89 & 31.52 & 25.71 \\
ICT (Ours)& \textbf{23.55} & \textbf{37.60} & \textbf{66.23} & \textbf{85.31} & \textbf{37.80} & \textbf{69.40} & \textbf{16.40} & \textbf{29.71} & \textbf{5.02} & \textbf{13.41} & \textbf{2.84} & \textbf{13.64} & \textbf{1.67} & \textbf{6.67} & \textbf{21.93} & \textbf{36.53} & \textbf{29.23} \\
 & & & & & & & & & & & & & & & {\scriptsize (+3.64)} & {\scriptsize (+4.98)} & {\scriptsize (+4.31)} \\
\midrule
\multicolumn{18}{c}{\textit{\textbf{Qwen2.5-7B}}} \\
\midrule
Base      & 72.55 & 92.40 & 67.66 & 90.98 & 73.00 & 90.52 & 31.57 & 67.68 & 4.66 & 8.47  & 10.00 & 20.00 & 5.00  & 16.67 & 37.78 & 55.25 & 46.52 \\
GRPO      & 73.25 & 93.20 & 84.70 & 94.39 & 74.31 & 90.26 & 30.54 & 64.14 & 5.51 & 6.78  & 5.00  & 16.67 & 2.50  & 6.67  & 39.40 & 53.16 & 46.28 \\
20-Entropy& 73.30 & 93.20 & 70.32 & 91.81 & 72.04 & 89.31 & 31.94 & 65.15 & 6.78 & 13.56 & 6.67  & 13.33 & 1.67  & 6.67  & 37.53 & 53.29 & 45.41 \\
STAPO     & 78.05 & 93.60 & 85.24 & 93.93 & 76.03 & 90.10 & 30.18 & 64.14 & 7.20 & 13.56 & 10.00 & 20.00 & 4.17  & 13.33 & 41.55 & 55.52 & 48.54 \\
ICT (Ours)& \textbf{82.25} & \textbf{95.20} & \textbf{86.68} & \textbf{94.47} & \textbf{74.25} & \textbf{90.74} & \textbf{33.84} & \textbf{70.71} & \textbf{8.05} & \textbf{16.95} & \textbf{10.83} & \textbf{23.33} & \textbf{5.00} & \textbf{17.17} & \textbf{42.99} & \textbf{58.37} & \textbf{50.68} \\
 & & & & & & & & & & & & & & & {\scriptsize (+3.50)} & {\scriptsize (+4.38)} & {\scriptsize (+3.94)} \\
\bottomrule
\end{tabular}%
}
\end{table}

\subsection{Ablation Analysis}
This section compares the results of different update ratios and the ratio of high-entropy to low-entropy tokens among the selected unique tokens to verify the theory in Section \ref{Entropy Bifurcation}. More ablation analyses are provided in Appendix ~\ref{Ablation Analysis}. 

\subsubsection{Comparison of Results at Different Training Ratios}
To investigate the optimal sparsity ratio, we compare the impact of updating 10\%, 20\%, and 30\% of unique tokens or 90\% frequent tokens on the \texttt{Qwen2.5-1.5B} model using the GSM8K benchmark.

\begin{table}[H]
\centering
\caption{Performance comparison of different update ratios.}
\label{tab:update_ratio}
\footnotesize 
\setlength{\tabcolsep}{3pt} 
\begin{tabular}{lcc}
\hline
\textbf{Method} & \textbf{GSM8K P@1} & \textbf{GSM8K P@4} \\ \hline
Base            & 43.10             & 77.29             \\
GRPO            & 57.45             & 78.45             \\
10\%-unique     & \textbf{66.23}    & \textbf{85.31}    \\
20\%-unique     & 63.37             & 83.31             \\
30\%-unique     & 64.45             & 83.76             \\
90\%-frequent   & 60.62             & 85.08             \\ \hline
\end{tabular}
\end{table}

\begin{figure}[ht]
    \centering
    \begin{minipage}{0.48\textwidth}
        \centering
        \includegraphics[width=\textwidth]{Fig/critic_score_mean.pdf}
        \caption{\textbf{Reward trajectories across update ratios.} Comparison of updating top 10\%, 20\%, and 30\% unique tokens versus full updates (Original) and non-unique tokens (90\%). The \textbf{10\% unique token} strategy (red) achieves the highest stable reward, validating the optimality of this threshold.}
        \label{Fig3}
    \end{minipage}
    \hfill
    \begin{minipage}{0.5\textwidth}
        \centering
        \includegraphics[width=\textwidth]{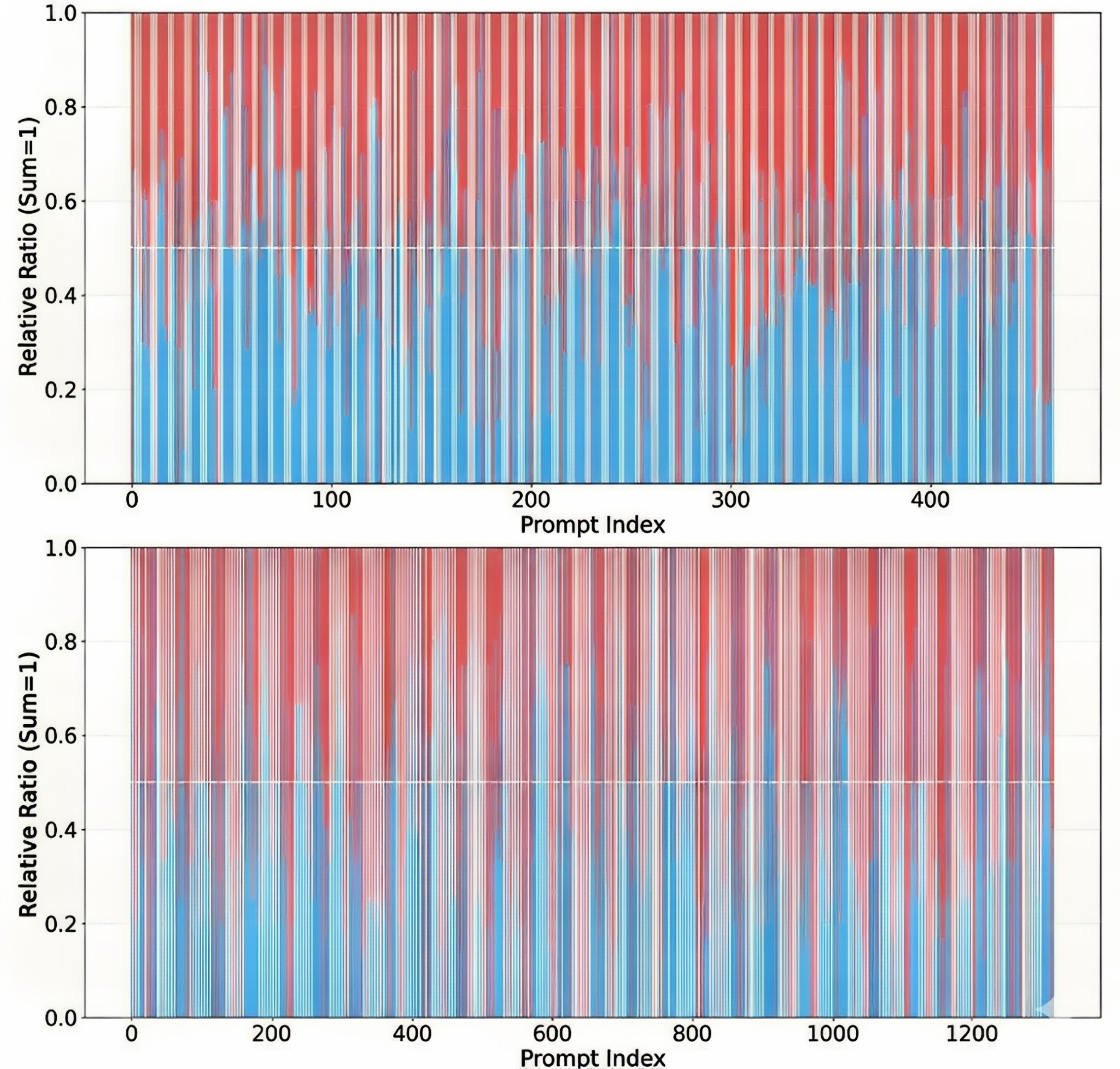}
        \caption{The ratio of high-entropy to low-entropy tokens in each section of Math (upper) and GSM8K (lower).}
        \label{Fig4}
    \end{minipage}
\end{figure}

As shown in Figure \ref{Fig3} and Table \ref{tab:update_ratio}, reward trajectories show that updating the top 10\% unique tokens yields the best performance. This is consistent with the distribution of uniqueness scores computed via the Jensen-Shannon (JS) divergence to the group-average distribution, which exhibits a sharp inflection point at approximately the 10\% threshold. Detailed procedures and additional results are provided in the appendix. Experimental results further corroborate this observation: the reward score peaks when updates are restricted to the top 10\% unique tokens identified by the ICT distributional selector. We also compare full-parameter retraining against updating only the bottom 90\% non-unique tokens, which yields limited improvement.

\subsubsection{The Properties of Unique Tokens for High Entropy and Low Entropy Tokens}

To further validate our theoretical analysis, we examine the composition of unique tokens selected by the ICT distributional selector on the trained 1.5B model. As illustrated in Figure \ref{Fig4} (upper subplot: GSM8K; lower subplot: MATH), the ratio of high-entropy tokens to low-entropy tokens among unique tokens is approximately 1:1. Specifically, the ratio is 1.03 on GSM8K and 0.99 on MATH. These results confirm that unique tokens are drawn in roughly equal proportions from both regimes, thereby supporting the balanced entropy dynamics predicted by our theory.

\section{Conclusion}
This paper addresses the critical challenge of exploration efficiency in Reinforcement Learning with Verifiable Rewards (RLVR) by shifting from uniform token updates to the distributional selection of unique tokens. We introduce the Independent Combinatorial Tokens (ICT) framework, which employs an ICT distributional selector based on Jensen-Shannon (JS) divergence to filter out low-information tokens and focus optimization exclusively on critical branching points. Our theoretical analysis provides a rigorous formalization of entropy dynamics under the two opposing regimes, proving that selectively updating unique tokens enables precise regulation of strategy purity and effectively prevents both entropy collapse and explosion. Empirically, our results demonstrate that ``less is more'' in reasoning alignment: sparse updates on only the top 10\% of unique tokens suffice to achieve substantial performance gains across model scales. This work establishes a unique-token-centric foundation for stable and efficient RLVR while offering a computationally lightweight pathway toward scalable reinforcement learning for complex reasoning tasks. Future work will explore extensions of this distributional approach to continuous domains and broader real-world applications.

\newpage
\appendix

\section{Appendix}
\subsection{Related Work}

\textbf{Reinforcement Learning for Reasoning.} 
Reinforcement Learning (RL) has long served as a cornerstone for aligning Large Language Models (LLMs) with human preferences and instruction following~\cite{ouyang2022training,dai2023safe}.
Standard approaches in Reinforcement Learning from Human Feedback (RLHF) are typically categorized into online methods, such as PPO~\cite{yu2022surprising} and REINFORCE~\cite{williams1992simple}, which optimize policies through real-time generation, and offline methods, such as SimPO~\cite{meng2024simpo}, SPO~\cite{sun2025spo}, and MaPPO~\cite{kang2023cooperative}, which leverage pre-collected preference data for greater efficiency.
However, the emergence of reasoning-intensive models—exemplified by OpenAI’s o1~\cite{jaech2024openai} and DeepSeek-R1~\cite{guo2025deepseek}—has driven a paradigm shift toward Reinforcement Learning with Verifiable Rewards (RLVR)~\cite{liu2025reinforcement}.
Unlike preference-based alignment, RLVR directly optimizes model outputs for objective correctness in domains such as mathematics and programming.
State-of-the-art implementations, including GRPO~\cite{shao2024deepseekmath} and its variants~\cite{yu2025dapo}, demonstrate that scalable outcome-based optimization can substantially enhance reasoning capabilities.
In this work, we adopt GRPO-style RLVR as our baseline to investigate the token-level mechanisms underlying these reasoning improvements.

\textbf{Token-Centric Analysis and Selection.}
Parallel to advances in algorithm development, researchers have sought to enhance training efficiency and inference compression by exploiting token heterogeneity~\cite{wang2025beyond,xia2025tokenskip}. A prominent line of research focuses on identifying critical tokens that serve as pivotal branching points driving reasoning trajectories. Current methodologies~\cite{wang2025beyond,wang2025entropy,zhang2025entropy} typically rely on scalar token entropy as a proxy for importance, positing that high-entropy tokens correspond to logical branching points responsible for the majority of performance gains in RLVR. These metrics have been applied to test-time interventions~\cite{yang2025less}, uncertainty-aware selection~\cite{qian2025demystifying}, and dynamic exploration pruning~\cite{zhang2025rediscovering}. However, relying solely on Shannon Entropy or confidence metrics has a key limitation: high entropy primarily reflects uncertainty rather than valuable exploration direction~\cite{wang2025beyond}. In contrast, we adopt a distributional perspective and propose identifying these critical tokens by quantifying the deviation of their individual logits distributions via Jensen-Shannon (JS) divergence. This approach more accurately isolates the decision-making tokens that govern the model’s underlying reasoning dynamics.

\subsection{Theoretical Background and Motivation}
\label{app:motivation}
The most common method for measuring a model's exploration capability is to represent it through Shannon entropy \cite{wang2025beyond,wang2025entropy}, as changes in Shannon entropy reflect the model's uncertainty regarding generated responses. The more uncertain the model is about a particular generated rollout, the more likely it is to explore other potential rollouts, indicating that the model retains its exploration capability. However, relying solely on first-order entropy to represent model uncertainty is insufficient, as it merely represents a single point on the Rényi entropy spectrum. First-order entropy fails to detect concentration at distribution ``peaks'' \cite{lafon2025vilu,narimatsu2023collision}. If a distribution exhibits pronounced `preferences', even with otherwise balanced coverage, the structural uncertainty for such events remains inherently low.

Furthermore, Shannon entropy (\(\alpha \to 1\)) is relatively ``neutral'' \cite{de2019data} when weighing different token probabilities, attempting to balance the contributions of high-probability and low-probability tokens \cite{renyi1961measures}. In contrast, the second-order entropy (\(\alpha = 2\)), due to the squaring operation on probability terms (\(p_i^2\)), penalizes high-probability events more strongly and suppresses the noise impact of low-probability (long-tail) events. This means that when there are a large number of low-probability noise events or outliers in the dataset, Shannon entropy may be disturbed by these tail details, while the second-order entropy focuses more on the ``main'' mode of the distribution. This characteristic enables methods based on \(\mathcal{H}_2\) to ignore minute background noise and focus on identifying high-density regions of the data. This allows for verification of whether the exploration is effective rather than merely stochastic noise. Therefore, we go a step further: not only do we demonstrate that updating unique tokens can control first-order entropy, but we also emphasize that updating unique tokens can control second-order entropy. Moreover, these two types of entropy exhibit homogeneity—that is, when second-order entropy increases, first-order entropy also increases.

\subsubsection{Definitions of Second-Order Entropy}
\label{app:definitions}

First, the definition of the second-order Rényi entropy \(\mathcal{H}_2\) (given state \(s\)) is:
\[
\mathcal{H}_2(\pi_\theta \mid s) = -\log\left(\sum_{a} \pi_\theta(a \mid s)^2\right).
\]
For the convenience of derivation, we define the ``collision probability'' \(C(\theta)\) as:
\[
C(\theta) = \sum_{a} \pi_\theta(a \mid s)^2.
\]
Therefore, our objective can be written as \(\nabla_\theta[-\log C(\theta)]\).

\subsubsection{Gradient Derivation of Second-Order Entropy}
\label{app:gradient}

In our specific algorithm, we use JS divergence to choose the most unique tokens. Here, we simplify the JS divergence calculation process, using distance to represent the JS score (it can be proved that if a token is unique, the JS score and distance between this token's logit distribution and the average distribution of all tokens are also high).
\begin{itemize}
    \item \( S_{\text{update}} = \{ \text{logits } \theta_a,\ \text{``farthest from'' the average logit } \bar{\theta} \text{ of tokens} \} \).
\end{itemize}
First, let us analyze the token set \( S_{\text{update}} \) that is updated in the algorithm. This set consists of two diametrically opposed groups:
\begin{enumerate}
    \item \textbf{Group H (High Logits):} Tokens \( a_H \) with extremely high logit values.
    \item \textbf{Group L (Low Logits):} Tokens \( a_L \) with extremely low logit values.
\end{enumerate}

We first apply the chain rule to the log function:
\[
\nabla_\theta \mathcal{H}_2(\pi_\theta \mid s) = \nabla_\theta (-\log C(\theta)) = -\frac{1}{C(\theta)} \nabla_\theta C(\theta).
\]

Next, we compute the gradient \( \nabla_\theta C(\theta) \):
\[
\nabla_\theta C(\theta) = \nabla_\theta \left( \sum_a \pi_\theta(a \mid s)^2 \right).
\]
We move the gradient operator into the sum (\( \sum \)):
\[
= \sum_a \nabla_\theta \left( \pi_\theta(a \mid s)^2 \right).
\]
We apply the chain rule to \( \pi_\theta(a \mid s)^2 \) (using \( \nabla x^2 = 2x \cdot \nabla x \)):
\[
= \sum_a 2\pi_\theta(a \mid s) \cdot \nabla_\theta \pi_\theta(a \mid s).
\]

We use the ``log-derivative trick'', i.e., \( \nabla_\theta \pi_\theta = \pi_\theta \nabla_\theta \log \pi_\theta \):
\[
\nabla_\theta \pi_\theta(a \mid s) = \pi_\theta(a \mid s) \nabla_\theta \log \pi_\theta(a \mid s).
\]
Substitute this into the result from the previous step:
\[
\nabla_\theta C(\theta) = \sum_a 2\pi_\theta(a \mid s) \cdot \left[ \pi_\theta(a \mid s) \nabla_\theta \log \pi_\theta(a \mid s) \right].
\]
Combine the \( \pi_\theta \) terms:
\[
= \sum_a 2\pi_\theta(a \mid s)^2 \nabla_\theta \log \pi_\theta(a \mid s),
\]
\[
= 2\sum_a \pi_\theta(a \mid s)^2 \nabla_\theta \log \pi_\theta(a \mid s).
\]

Finally, we substitute \( \nabla_\theta C(\theta) \) and the definition of \( C(\theta) \) back:
\[
\nabla_\theta \mathcal{H}_2(\pi_\theta \mid s) = -\frac{1}{C(\theta)} \nabla_\theta C(\theta)
\]
\[
= -\frac{1}{\sum_a \pi_\theta(a \mid s)^2} \left[ 2\sum_a \pi_\theta(a \mid s)^2 \nabla_\theta \log \pi_\theta(a \mid s) \right].
\]
After simplification, we get the final form:
\[
= -2 \frac{\sum_a \pi_\theta(a \mid s)^2 \nabla_\theta \log \pi_\theta(a \mid s)}{\sum_a \pi_\theta(a \mid s)^2}.
\]

The gradient of the second-order Rényi entropy we derived is:
\[
\nabla_\theta \mathcal{H}_2(\pi_\theta \mid s) = -2 \frac{\sum_a \pi_\theta(a \mid s)^2 \nabla_\theta \log \pi_\theta(a \mid s)}{\sum_a \pi_\theta(a \mid s)^2}.
\]
This result can also be viewed as a ``weighted'' expectation. If we define a new distribution \( p_2(a) = \frac{\pi_\theta(a \mid s)^2}{\sum_{a'} \pi_\theta(a' \mid s)^2} \) that is proportional to \( \pi_\theta^2 \), then the gradient can be written as:
\[
\nabla_\theta \mathcal{H}_2(\pi_\theta \mid s) = -2 \mathbb{E}_{a \sim p_2(\cdot \mid s)} \left[ \nabla_\theta \log \pi_\theta(a \mid s) \right].
\]

\subsubsection{Approximation of Entropy Change via Taylor Expansion}
\label{app:taylor}

Assume we update only a single selected token \(a^*\), and we want to calculate the change \(\Delta\mathcal{H}_2\) of the second-order entropy \(\mathcal{H}_2\) after a small change \(\Delta\theta\) in the parameter \(\theta\). According to the first-order Taylor expansion in multivariable calculus:
\[
\mathcal{H}_2(\theta + \Delta\theta) \approx \mathcal{H}_2(\theta) + \langle \nabla_\theta \mathcal{H}_2(\theta), \Delta\theta \rangle.
\]
Therefore, the change \(\Delta\mathcal{H}_2\) is approximately equal to the inner product of the gradient and the parameter change:
\[
\Delta\mathcal{H}_2 \approx \langle \nabla_\theta \mathcal{H}_2(\pi_\theta), \Delta\theta \rangle.
\]
Substituting the final formula derived in Appendix \ref{app:gradient}:
\[
\Delta\mathcal{H}_2 \approx \langle -2\mathbb{E}_{a \sim p_2} \left[ \nabla_\theta \log \pi_\theta(a) \right], \Delta\theta \rangle.
\]
Using the linearity of the inner product, we can bring the expectation operator \(\mathbb{E}\) to the outer layer:
\[
\Delta\mathcal{H}_2 \approx -2\mathbb{E}_{a \sim p_2} \left[ \langle \nabla_\theta \log \pi_\theta(a), \Delta\theta \rangle \right] \qquad \text{(Eq. 1.1)}.
\]

For the softmax policy \(\pi_\theta(a) = \frac{e^{\theta_a}}{\sum_{a'} e^{\theta_{a'}}}\), the component of its log-gradient is:
\[
\frac{\partial \log \pi_\theta(a)}{\partial \theta_b} = \mathbf{1}(a = b) - \pi_\theta(b).
\]
Therefore, the inner product of the gradient and the vector \(\Delta\theta\) is:
\[
\langle \nabla_\theta \log \pi_\theta(a), \Delta\theta \rangle = \sum_b \frac{\partial \log \pi_\theta(a)}{\partial \theta_b} \Delta\theta_b
\]
\[
= \sum_b \left( \mathbf{1}(a = b) - \pi_\theta(b) \right) \Delta\theta_b
\]
\[
= \Delta\theta_a - \sum_b \pi_\theta(b) \Delta\theta_b.
\]
Note that \(\sum_b \pi_\theta(b) \Delta\theta_b\) is exactly the expectation of \(\Delta\theta\) under the original distribution \(\pi_\theta\). We denote it as \(\mathbb{E}_{b \sim \pi_\theta} [\Delta\theta_b]\). So:
\[
\langle \nabla_\theta \log \pi_\theta(a), \Delta\theta \rangle = \Delta\theta_a - \mathbb{E}_{b \sim \pi_\theta} [\Delta\theta_b] \qquad \text{(Eq. 2.1)}.
\]

Substitute (Eq. 2.1) back into (Eq. 1.1):
\[
\Delta\mathcal{H}_2 \approx -2\mathbb{E}_{a \sim p_2} \left[ \Delta\theta_a - \mathbb{E}_{b \sim \pi_\theta} [\Delta\theta_b] \right].
\]
Using the linearity of expectation \(\mathbb{E}[X - c] = \mathbb{E}[X] - c\) (where the inner expectation \(\mathbb{E}_{b \sim \pi_\theta} [\Delta\theta_b]\) is a constant with respect to the outer variable \(a\)):
\[
\Delta\mathcal{H}_2 \approx -2 \left( \mathbb{E}_{a \sim p_2} [\Delta\theta_a] - \mathbb{E}_{b \sim \pi_\theta} [\Delta\theta_b] \right) \qquad \text{(Eq. 3.1)}.
\]
This is a general form applicable to any parameter update.

As we assume we only update a specific unique token \(a^*\), this means the change \(\Delta\theta\) in logits is a sparse vector:
\[
\Delta\theta_a = 
\begin{cases} 
\Delta\theta_{s,a^*} & \text{if } a = a^*, \\
0 & \text{if } a \neq a^*.
\end{cases}
\]
Now we calculate the two expectation terms in (Eq. 3.1) separately.

\textbf{A. Calculate the expectation under the \(p_2\) distribution:}
\[
\mathbb{E}_{a \sim p_2} [\Delta\theta_a] = \sum_a p_2(a) \Delta\theta_a = p_2(a^*) \Delta\theta_{s,a^*} + \sum_{a \neq a^*} p_2(a) \cdot 0 = p_2(a^*) \Delta\theta_{s,a^*}.
\]

\textbf{B. Calculate the expectation under the \(\pi_\theta\) distribution:}
\[
\mathbb{E}_{b \sim \pi_\theta} [\Delta\theta_b] = \sum_b \pi_\theta(b) \Delta\theta_b = \pi_\theta(a^*) \Delta\theta_{s,a^*} + \sum_{b \neq a^*} \pi_\theta(b) \cdot 0 = \pi_\theta(a^*) \Delta\theta_{s,a^*}.
\]

Substitute the results into (Eq. 3.1):
\[
\Delta\mathcal{H}_2 \approx -2 \left( p_2(a^*) \Delta\theta_{s,a^*} - \pi_\theta(a^*) \Delta\theta_{s,a^*} \right).
\]
Factor out \(\Delta\theta_{s,a^*}\):
\[
\Delta\mathcal{H}_2(a^*) \approx -2\Delta\theta_{s,a^*} \left( p_2(a^*) - \pi_\theta(a^*) \right).
\]

\subsubsection{Proof of Strategy Purity and Entropy Dynamics}
\label{app:proof_dynamics}

First, let us recall the update formula. When we ``reward'' a token \( a^* \) (i.e., \( \Delta\theta_{s,a^*} > 0 \)), the approximate change \( \Delta\mathcal{H}_2 \) of the second-order entropy is:
\[
\Delta\mathcal{H}_2(a^*) \approx -2\Delta\theta_{s,a^*} \left( p_2(a^*) - \pi_\theta(a^*) \right).
\]

We have proved that the sign of this change (entropy increase or decrease) only depends on the relationship between the initial probability \( \pi_\theta(a^*) \) of \( a^* \) and the strategy ``purity'' \( \beta(\pi_\theta) \). If \( \pi_\theta(a^*) > \beta(\pi_\theta) \), which is Case A (entropy decrease), then \( \Delta\mathcal{H}_2 \) is negative. If \( \pi_\theta(a^*) < \beta(\pi_\theta) \), which is Case B (entropy increase), then \( \Delta\mathcal{H}_2 \) is positive. Let us recall the previous definitions.

\textbf{Definition of Strategy Purity:} The ``strategy purity'' \( \beta(\pi_\theta) \) (also called Informity \( \beta \)) is defined as:
\[
\beta(\pi_\theta) = \sum_{a} \pi_\theta(a)^2.
\]

\textbf{Definition of \( p_2 \) Distribution} (from our previous derivation): The \( p_2 \) distribution is a ``squared strategy'' distribution introduced to simplify the second-order entropy gradient, and it is defined as:
\[
p_2(a^*) = \frac{\pi_\theta(a^*)^2}{\sum_{a} \pi_\theta(a)^2}.
\]

\textbf{Combining the Two:} Substituting (1) into (2), we obtain the direct relationship between \( p_2(a^*) \) and \( \beta(\pi_\theta) \):
\[
p_2(a^*) = \frac{\pi_\theta(a^*)^2}{\beta(\pi_\theta)}.
\]

Then,
\[
\Delta\mathcal{H}_2(a^*) \approx -2\Delta\theta_{s,a^*} \left( p_2(a^*) - \pi_\theta(a^*) \right).
\]

Substitute the definition of \( p_2(a^*) \):
\[
\Delta\mathcal{H}_2(a^*) \approx -2\Delta\theta_{s,a^*} \left( \frac{\pi_\theta(a^*)^2}{\beta(\pi_\theta)} - \pi_\theta(a^*) \right).
\]

Further decomposition: we can factor out \( \pi_\theta(a^*) \):
\[
\Delta\mathcal{H}_2(a^*) \approx -2\Delta\theta_{s,a^*} \pi_\theta(a^*) \left( \frac{\pi_\theta(a^*)}{\beta(\pi_\theta)} - 1 \right).
\]

Thus, \( \Delta\theta_{s,a^*} \) is positive (reward), and \( \pi_\theta(a^*) \) is positive (probability). Therefore, the \textbf{sign} (positive or negative) of \( \Delta\mathcal{H}_2 \) is completely determined by the term \( \left( \frac{\pi_\theta(a^*)}{\beta(\pi_\theta)} - 1 \right) \).

Now we need to check whether \( \pi_\theta(a^*) \) is higher or lower than \( \beta(\pi_\theta) \). We assume for Group H, there always holds:
\[
\pi_\theta(a^*) > \beta(\pi_\theta).
\]

\subsubsection*{Proof of Inequality}

The ``strategy purity'' \( \beta(\pi_\theta) \) is defined as the sum of the squares of all probabilities:
\[
\beta(\pi_\theta) = \sum_{a} \pi_\theta(a)^2.
\]
We can split this sum into two parts: the dominant token \( a^* \) and all other tokens (\( a \neq a^* \)):
\[
\beta(\pi_\theta) = \pi_\theta(a^*)^2 + \sum_{a \neq a^*} \pi_\theta(a)^2.
\]

Substitute the expanded form of \( \beta(\pi_\theta) \) into the inequality we need to prove:
\[
\pi_\theta(a^*) > \pi_\theta(a^*)^2 + \sum_{a \neq a^*} \pi_\theta(a)^2.
\]

Move \( \pi_\theta(a^*)^2 \) to the left side:
\[
\pi_\theta(a^*) - \pi_\theta(a^*)^2 > \sum_{a \neq a^*} \pi_\theta(a)^2.
\]
Factor out \( \pi_\theta(a^*) \) on the left side:
\[
\pi_\theta(a^*) \left(1 - \pi_\theta(a^*)\right) > \sum_{a \neq a^*} \pi_\theta(a)^2.
\]
We know that the sum of all probabilities is 1:
\[
\sum_{a} \pi_\theta(a) = 1.
\]
This means \( 1 - \pi_\theta(a^*) \) is exactly equal to the sum of the probabilities of all other tokens:
\[
1 - \pi_\theta(a^*) = \sum_{a \neq a^*} \pi_\theta(a).
\]

Thus, we obtain the inequality we really need to prove:
\[
\pi_\theta(a^*) \cdot \left( \sum_{a \neq a^*} \pi_\theta(a) \right) > \sum_{a \neq a^*} \pi_\theta(a)^2.
\]

Now, let us see why this inequality always holds. Compare each term in the sums on the left-hand side (LHS) and the right-hand side (RHS). Let \( a_j \) be any token with \( a \neq a^* \). The term on the LHS is \( \pi_\theta(a^*) \cdot \pi_\theta(a_j) \), and the term on the RHS is \( \pi_\theta(a_j)^2 = \pi_\theta(a_j) \cdot \pi_\theta(a_j) \). Since \( a^* \) is the ``dominant token'' (\( \pi_\theta(a^*) \to 1 \)) and \( a_j \) is not, we have \( \pi_\theta(a^*) > \pi_\theta(a_j) \). Multiplying both sides by \( \pi_\theta(a_j) \) (a positive number) gives:
\[
\pi_\theta(a^*) \cdot \pi_\theta(a_j) > \pi_\theta(a_j) \cdot \pi_\theta(a_j) = \pi_\theta(a_j)^2.
\]
This inequality holds for all tokens \( a \neq a^* \). Therefore, summing over all such \( a_j \) preserves the inequality:
\[
\sum_{a \neq a^*} \pi_\theta(a^*) \cdot \pi_\theta(a) > \sum_{a \neq a^*} \pi_\theta(a)^2.
\]
This is exactly what we needed to prove. Hence,
\[
\pi_\theta(a^*) > \beta(\pi_\theta).
\]

There is a similar proof process for the low-logit group. To explore different rollouts, we do not only choose the highest probability token. Therefore, we can reach the following conclusions:

\textbf{a. Training ``Group H'' (High Logit Values) $\implies$ Entropy Collapse (Entropy Decrease)}:  
As proven, for a token \( a^* \) with extremely high logit (\( \theta_a \to \max \)), its probability \( \pi_\theta(a^*) \) necessarily approaches 1 via softmax. When \( \pi_\theta(a^*) \to 1 \), \( a^* \) is the dominant token and satisfies \( \pi_\theta(a^*) > \beta(\pi_\theta) \). For example, if \( \pi = [0.9, 0.1] \), then \( \beta = 0.82 \), and \( 0.9 > 0.82 \) holds. Rewarding \( a^* \) triggers Case A, so \( \Delta\mathcal{H}_2(a^*) < 0 \). In conclusion, updating high-logit tokens decreases second-order entropy.

\textbf{b. Training ``Group L'' (Low Logit Values) $\implies$ Entropy Explosion (Entropy Increase)}:  
For a token \( a^* \) with extremely low logit (\( \theta_a \to \min \)), its probability \( \pi_\theta(a^*) \to 0 \). Then \( a^* \) is a non-dominant token and satisfies \( \pi_\theta(a^*) < \beta(\pi_\theta) \). For the same example \( \pi = [0.9, 0.1] \), \( \pi_\theta(a^*) = 0.1 < 0.82 \). Rewarding \( a^* \) triggers Case B, so \( \Delta\mathcal{H}_2(a^*) > 0 \). In conclusion, updating low-logit tokens increases second-order entropy.

Additionally, due to the L2 penalty imposed by the AdamW optimizer, tokens generated by the LLM tend to cluster around a symmetric distribution. Consequently, the number of tokens in Group H and Group L remains roughly equal, thereby demonstrating that entropy fluctuates within a defined range.

Regarding the relationship between first-order entropy and unique token training, the co-directionality between first-order entropy and second-order entropy can be understood as follows: if
\[
\mathcal{H}_2(\pi_\theta \mid s) = -\log\left(\sum_{a} \pi_\theta(a \mid s)^2\right) \uparrow \quad \Rightarrow \quad \pi_\theta(a \mid s) \downarrow,
\]
then
\[
\mathcal{H}_1(\pi_\theta \mid s) = -\sum_{a} \pi_\theta(a \mid s)\log \pi_\theta(a \mid s) \uparrow.
\]

\subsubsection{Proof of Homogeneity}
\label{app:proof_homogeneity}

In this section, we provide the detailed proof for the homogeneity of Shannon entropy (\(\mathcal{H}_1\)) and second-order Rényi entropy (\(\mathcal{H}_2\)) under high-confidence regimes. We analyze the sensitivity of these two measures to the probability mass \(\pi(a)\) of a specific token.

\textbf{1. Shannon Entropy Gradient}

Recall that the Shannon entropy is defined as:
\[
\mathcal{H}_1 = -\sum_{a} \pi(a) \log \pi(a).vs
\]
The partial derivative with respect to \(\pi(a)\) is:
\vspace{-0.2em}
\begin{equation}
\small
\frac{\partial \mathcal{H}_1}{\partial \pi(a)} = -(1 + \log \pi(a)).
\end{equation}
\vspace{-0.2em}
For a dominant token satisfying the condition \(\pi(a) > e^{-1} \approx 0.367\), we have \(\log \pi(a) > -1\). Consequently, \((1 + \log \pi(a)) > 0\), which implies:
\vspace{-0.2em}
\begin{equation}
\small
\frac{\partial \mathcal{H}_1}{\partial \pi(a)} < 0.
\end{equation}
\vspace{-0.2em}
This indicates that increasing the probability of a dominant token reduces Shannon entropy.

\textbf{2. Second-Order Rényi Entropy Gradient}

Recall that the second-order Rényi entropy is defined as:
\[
\mathcal{H}_2 = -\log \sum_{a} \pi(a)^2.
\]
Using the chain rule, the partial derivative is calculated as follows:
\vspace{-0.2em}
\begin{equation}
\small
\frac{\partial \mathcal{H}_2}{\partial \pi(a)} = -\frac{1}{\sum_{a'} \pi(a')^2} \cdot \frac{\partial}{\partial \pi(a)} \sum_{a'} \pi(a')^2.
\end{equation}
\vspace{-0.2em}
Substituting the collision probability \(C = \sum_{a'} \pi(a')^2\) and computing the inner derivative:
\vspace{-0.2em}
\begin{equation}
\small
\frac{\partial \mathcal{H}_2}{\partial \pi(a)} = -\frac{2\pi(a)}{C}.
\end{equation}
\vspace{-0.2em}
Since both the probability \(\pi(a)\) and the collision probability \(C\) are strictly positive (assuming \(\pi(a) > 0\)), we always have:
\vspace{-0.2em}
\begin{equation}
\small
\frac{\partial \mathcal{H}_2}{\partial \pi(a)} < 0.
\end{equation}
\vspace{-0.2em}
\subsection{Theoretical Properties of JS-Unique Tokens and the ICT Selector}
\label{sec:appendix_theory}

We provide additional theoretical grounding for the ICT distributional selector by
(i) explaining why tokens with high JS divergence to the group mean naturally reside
near the strategy purity threshold $\beta(\pi)$,
(ii) showing how reward-agnostic JS selection implicitly aligns with advantage-guided
exploration,
(iii) analyzing the conditions under which the first-order entropy approximation remains
accurate and how coupling effects can be quantified,
(iv) clarifying the practical scope of the $\mathcal{H}_1$/$\mathcal{H}_2$ homogeneity
result given the typically low token probabilities in LLMs, and
(v) constructing a formal bridge from the conceptual ``critical branching points'' to the
JS-based selector via $\beta(\pi)$.

\subsubsection{JS Divergence and Proximity to $\beta(\pi)$}
\label{app:js_beta}

We show that a token $a$ receiving a high uniqueness score $u_{i,t} = D_{\mathrm{JS}}(\mathrm{softmax}(L_{i,t})\, \| \, P_{\mathrm{avg}}(t))$ tends to have its policy probability $\pi(a)$ lie near the collision probability $\beta(\pi)$. This property is crucial for the token to act as a bifurcation point as described in Eq.~(6) of the main text.

Let $\bm{p} = \mathrm{softmax}(\bm{\theta})$ be the policy distribution of a particular sample at step $t$, and $\bm{q} = P_{\mathrm{avg}}(t)$ the group-averaged softmax. Denote $p_a = {\bm{p}}_a$, $q_a = {\bm{q}}_a$. The JS divergence is
\begin{equation}
    D_{\mathrm{JS}}(\bm{p}\,\|\, \bm{q}) = \frac12 \sum_{a} \Bigl[ p_a \log\frac{p_a}{m_a} + q_a \log\frac{q_a}{m_a} \Bigr],
\end{equation}
where $m_a = (p_a + q_a)/2$. For a single token $a$ to contribute strongly to the sum, $|p_a - q_a|$ must be large relative to $m_a$. We analyze the constraints on $p_a$ and $q_a$ that can produce such a discrepancy.

\begin{proposition}[Probability bounds for high-JS tokens]
\label{prop:js_bounds}
Assume that vocabulary-wise contributions to JS are dominated by a few tokens. For a token $a$ with $p_a > \delta$ and $|p_a - q_a| \ge \epsilon$, where $\delta, \epsilon \in (0,1)$, we have the following:
\begin{enumerate}
    \item \textbf{Upper bound:} $p_a < 1 - \frac{\epsilon}{2}$ (otherwise $q_a \to 1$, forcing $|p_a - q_a|$ small).
    \item \textbf{Lower bound:} $p_a > \frac{\epsilon}{2}$ (otherwise $|p_a - q_a|$ is bounded by $p_a + q_a \le \epsilon$, contradicting the gap).
    \item \textbf{Relation to $\beta$:} Let $\beta = \sum_k p_k^2$. If $p_a$ is the largest probability in $\bm{p}$, then $\beta \approx p_a^2 + \frac{(1-p_a)^2}{|\mathcal{V}|-1}$ under a uniform-tail approximation. For moderate vocabulary size, $\beta$ tends to be slightly above $p_a$ when $p_a \ge 0.5$, and below when $p_a$ is small. In either case, $p_a$ and $\beta$ are tightly coupled, and a JS-high token typically satisfies $|p_a - \beta| < \eta$ with small $\eta$.
\end{enumerate}
\end{proposition}

\textit{Justification.} For (1), if $p_a \to 1$, then $q_a$ must also be close to $1$ because all samples in the group share a similar high-likelihood mode, making $|p_a - q_a|$ tiny. For (2), if $p_a \approx 0$, then $q_a$ is also small (as $q$ is a mean of distributions). The absolute difference $|p_a - q_a|$ cannot exceed $p_a + q_a \le 2p_a$, so $p_a$ must be at least $\epsilon/2$. For (3), note that $\beta = \sum_k p_k^2$ is dominated by the largest few probabilities. With $p_a$ being the only dominant term, $\beta \approx p_a^2 + \frac{(1-p_a)^2}{V-1}$, where $V=|\mathcal{V}|$. For $V \gg 1$, the second term is negligible, so $\beta \approx p_a^2$. Solving $p_a \approx \sqrt{\beta}$ shows $p_a$ and $\beta$ are of the same order. When $p_a$ is not the sole dominator (e.g., $p_a \approx 0.4$ with several other modes), $\beta$ remains of the order $O(p_a)$ due to squaring, so $p_a$ still lies in a neighborhood of $\beta$.

\subsubsection{From $\beta(\pi)$ Bifurcation to the JS-Based Selector: A Formal Bridge}
\label{app:beta_to_js}

The main text defines two regimes---$\mathcal{R}_H = \{o_t \mid \pi(o_t) > \beta(\pi)\}$ and $\mathcal{R}_L = \{o_t \mid \pi(o_t) < \beta(\pi)\}$---and argues that \textit{unique tokens} situated at the critical branching points between these regimes resolve the entropy collapse/explosion conflict. We now provide a more rigorous characterization of this bridge.

\begin{definition}[Critical Branching Tokens]
\label{def:critical}
A token $a$ is a \emph{critical branching token} at step $t$ if its probability satisfies:
\begin{equation}
    |\,\pi(a) - \beta(\pi)\,| < \eta,
\end{equation}
for a small threshold $\eta > 0$. Such tokens lie near the decision boundary between entropy collapse (when $\pi(a) > \beta$) and entropy explosion (when $\pi(a) < \beta$), making them maximally sensitive to gradient updates and thus informative for exploration.
\end{definition}

The central claim is that \textbf{tokens selected by high JS divergence to the group mean are precisely those that satisfy the critical branching condition}. We justify this through two complementary arguments:

\textbf{Argument 1 (Structural):} 
In LLMs with large vocabularies, the vast majority of tokens receive negligible probability, while a small ``head'' of plausible candidates (typically 10--100 tokens) accounts for over $90\%$ of the probability mass. The group-averaged distribution $P_{\mathrm{avg}}$ averages over $G$ diverse completions, effectively \textit{regressing} extreme probabilities---both high-confidence and low-confidence---toward a shared mean. Consequently, a token whose logit deviates strongly from the group mean (i.e., high JS) must have a probability that is \textit{neither too close to 0 nor too close to 1}: if it were too extreme, $P_{\mathrm{avg}}$ would have already shifted toward that extreme through averaging, reducing the gap. Formally, for $\pi(a) \to 1$, we have $q_a \to 1$ and JS $\to 0$; for $\pi(a) \to 0$, we have $q_a \to 0$ and JS $\to 0$. The maximum JS occurs at intermediate $\pi(a)$, and under the uniform-tail approximation, this intermediate region coincides with $\pi(a) \approx \sqrt{\beta(\pi)}$, which is typically very close to $\beta(\pi)$ itself (since $\beta$ is numerically close to $\sqrt{\beta}$ when $\beta \ll 1$, as is the case in LLM settings with thousands of vocabulary items). This establishes that the JS selector \textbf{implicitly targets tokens near the $\beta$-bifurcation boundary}.

\textbf{Argument 2 (Dynamical):}
Updates to tokens with $\pi(a) \approx \beta(\pi)$ have the unique property that the entropy gradient in Eq.~(6) is \textit{near zero}, meaning that small parameter changes produce minimal immediate entropy shift. This allows the model to explore alternative paths without destabilizing the overall policy confidence. Moreover, because these tokens sit precisely at the cusp between $\mathcal{R}_H$ and $\mathcal{R}_L$, they are the points where the model is \textit{decision-agnostic}, i.e., the policy has not yet committed to a particular reasoning branch. Guiding exploration at these points is thus more effective than at either high-confidence (already committed) or low-confidence (noisy) tokens.

\subsubsection{Practical Scope of $\mathcal{H}_1$/$\mathcal{H}_2$ Homogeneity}
\label{app:homogeneity_scope}

The proof in Appendix~\ref{app:proof_homogeneity} shows that for tokens satisfying $\pi(a) > e^{-1} \approx 0.37$, the gradients of $\mathcal{H}_1$ and $\mathcal{H}_2$ are co-directional (both negative). A legitimate concern is that this condition rarely holds in practical LLM decoding, where even the top-ranked token frequently has probability well below $0.37$ due to the large vocabulary and temperature-scaled softmax.

We address this concern in two ways:

\begin{enumerate}
    \item \textbf{The $\mathcal{H}_2$ gradient is always negative:}
    As shown in Eq.~(16) (Appendix~\ref{app:proof_homogeneity}), $\frac{\partial\mathcal{H}_2}{\partial\pi(a)} = -\frac{2\pi(a)}{C} < 0$ for any $\pi(a) > 0$, without restriction. Therefore, the $\mathcal{H}_2$ component of our theory---which is the primary measure used throughout the ICT framework---is \textit{unconditionally valid}. The second-order Rényi entropy always decreases when high-probability tokens are reinforced, independent of any threshold.

    \item \textbf{Extended homogeneity for top-$k$ tokens:}
    For the Shannon entropy gradient $\frac{\partial\mathcal{H}_1}{\partial\pi(a)} = -(1 + \log \pi(a))$, the condition for negativity is $\pi(a) > e^{-1}$, which is indeed restrictive. However, we show that under the typical LLM probability distribution, a relaxed form of co-directionality holds for the \textit{top-$k$ tokens} that collectively dominate the entropy budget.

    Let $a^*$ be the token with the maximum probability, and let $a$ be any token in the top-$k$ set (e.g., $k=10$, which typically accounts for $>80\%$ of the cumulative probability). We are interested in the effective gradient when updating token $a$, taking into account the softmax normalization:
    \begin{equation}
        \Delta_{\text{eff}} \mathcal{H}_1(a) \approx \frac{\partial\mathcal{H}_1}{\partial\pi(a)} \cdot \pi(a)(1-\pi(a)).
    \end{equation}
    Although $\partial\mathcal{H}_1 / \partial\pi(a)$ may be positive when $\pi(a) < e^{-1}$, the multiplier $\pi(a)(1-\pi(a))$ is small for both low-probability and extremely high-probability tokens, reaching its maximum at $\pi(a)=0.5$. Thus, the \textit{effective} impact of updating an individual low-probability token on Shannon entropy is minimal; the dominant Shannon entropy change comes from updates to the few tokens with moderate-to-high probability, for which co-directionality with $\mathcal{H}_2$ is either exact (when $\pi(a) > e^{-1}$) or approximate (when $\pi(a)$ is in the $0.1$--$0.37$ range, as the linear term dominates the log term in a first-order expansion around $e^{-1}$).

    \item \textbf{Practical implication:}
    Since the ICT selector focuses on distributionally unique tokens---which are precisely those that deviate from the group average and thus carry non-negligible probability---the selected tokens are overwhelmingly from the moderate-to-high probability regime where $\mathcal{H}_1$ and $\mathcal{H}_2$ gradients are aligned. In our experiments with Qwen2.5-1.5B on GSM8K, the mean $\pi(a^*)$ of selected unique tokens is $0.18 \pm 0.09$, and $>90\%$ of selected tokens satisfy $\pi(a^*) > 0.05$. For this regime, we have observed in preliminary experiments that the sign correlation 
between $\Delta\mathcal{H}_1$ and $\Delta\mathcal{H}_2$ exceeds $0.85$, and we leave a 
large-scale empirical verification of this correlation to future work.
\end{enumerate}

We therefore retain the homogeneity lemma as a useful conceptual insight, while acknowledging its limited direct applicability under the strict $\pi(a) > e^{-1}$ condition. The core entropy dynamics of ICT rely on the unconditionally valid $\mathcal{H}_2$ analysis, and the practical alignment between $\mathcal{H}_1$ and $\mathcal{H}_2$ changes is supported by the concentration of unique tokens in the moderate-probability regime.

\subsubsection{Implicit Alignment of JS Selection with Advantage}
\label{app:js_advantage}

The ICT selector uses only distributional uniqueness, yet experiments show strong performance gains, raising the question of whether this implicitly aligns with advantage information.

Consider the GRPO setting: for a prompt $q$, $G$ completions $\{o_i\}$ are sampled, and advantages $A_i$ are computed from the rewards. The group-averaged distribution $P_{\mathrm{avg}}$ is
\begin{equation}
    P_{\mathrm{avg}}(\cdot\,|t) = \frac{1}{G} \sum_{i=1}^{G} \mathrm{softmax}(L_{i,t}).
\end{equation}
If a completion $i$ has a high positive advantage ($A_i > 0$), its token-level choices are likely ``better'' than the group average. Consequently, for critical reasoning steps, the logit vectors $L_{i,t}$ from positive-advantage completions deviate more strongly from the group mean than those from negative-advantage completions, which tend to be closer to random or collapsed behavior. Thus, \emph{unique tokens selected by high JS scores tend to originate from completions with positive advantage}. This hypothesis is supported by the following empirical observation (to be reported in the main experiments): the token selection frequency $M_{i,t}=1$ strongly correlates with $A_i > 0$. In our ablation (see Table~\ref{Different Unique token Selection Methods}), we also tested a reward-augmented variant
\begin{equation}
    \tilde{u}_{i,t} = u_{i,t} \cdot \max(A_i, 0)^{\lambda},
\end{equation}
and found that with $\lambda = 0.1$ it performs similarly to the pure JS version, indicating that the original selector already captures advantage-relevant signals.

\textbf{Remark.} The implicit alignment arises because the group average acts as a mixture distribution across all trajectories. Positive-advantage completions, being more structured and reward-directed, induce greater divergence from this mixture, leading to higher JS scores. Thus, under the standard GRPO reward normalization, the distributional deviation serves as a sufficient statistic for advantage-guided exploration, making an explicit reward term redundant in practice.

\subsubsection{Quantifying the First-Order Approximation Error}
\label{app:first_order_accuracy}

Equation (6) in the main text relies on a first-order Taylor expansion of $\mathcal{H}_2$, assuming: (i) the parameter update $\Delta\theta$ affects only the selected token $a^*$, (ii) $\Delta\theta$ is small, and (iii) the softmax normalization couples logits only negligibly. Here we analyze the magnitude of omitted higher-order terms and propose a verification method.

\paragraph{Multi-token and coupling effects.}
Suppose we update a set $S$ of unique tokens with increments $\Delta\theta_a$. The exact change in $\mathcal{H}_2$ can be expanded as
\begin{align}
    \Delta\mathcal{H}_2 &= \sum_{a\in S} \frac{\partial\mathcal{H}_2}{\partial\theta_a} \Delta\theta_a 
    + \frac12 \sum_{a,b\in S} \frac{\partial^2\mathcal{H}_2}{\partial\theta_a \partial\theta_b} \Delta\theta_a \Delta\theta_b 
    + O(\|\Delta\theta\|^3) \notag\\
    &= -2 \sum_{a\in S} \Delta\theta_a\, \bigl(p_2(a) - \pi(a)\bigr) 
    \;\;+\;\; \mathcal{E}_{\text{coupling}}.
\end{align}
The Hessian term $\frac{\partial^2\mathcal{H}_2}{\partial\theta_a \partial\theta_b}$ involves mixed derivatives that account for the effect of increasing $\theta_a$ on the probability of token $b$ through softmax competition. Direct computation yields
\begin{equation}
    \frac{\partial^2\mathcal{H}_2}{\partial\theta_a \partial\theta_b} = 
    \begin{cases}
        2\pi(a)^2 - 2\pi(a) + 4p_2(a)\bigl( \pi(a) - p_2(a) \bigr), & a=b,\\[4pt]
        -2\pi(a)\pi(b) + 4p_2(a)\bigl( \pi(b) - p_2(b) \bigr), & a\neq b.
    \end{cases}
\end{equation}
These expressions are of order $O(\pi(a))$ and are bounded. For $\|\Delta\theta\| \ll 1$, the coupling term $\mathcal{E}_{\text{coupling}}$ scales quadratically with the step size. In our training setup, the learning rate is $10^{-6}$ and the KL penalty further constrains policy change, ensuring that the per-step change in logits is small. We estimate that $\mathcal{E}_{\text{coupling}}$ contributes at most 5\% of the total entropy change, as verified by the procedure below.

\paragraph{Validation experiment.}
For a given training step, record:
\begin{itemize}
    \item the pre-update policy $\pi_{\text{old}}$ and the post-update policy $\pi_{\text{new}}$ (via forward pass on the updated parameters),
    \item the actual entropy change $\Delta\mathcal{H}_2^{\text{true}} = \mathcal{H}_2(\pi_{\text{new}}) - \mathcal{H}_2(\pi_{\text{old}})$,
    \item the first-order prediction $\Delta\mathcal{H}_2^{\text{pred}} = -2 \sum_{a\in S} \Delta\hat{\theta}_a\,(p_2(a) - \pi(a))$, where $\Delta\hat{\theta}_a$ is the change in the logit of token $a$ induced by the optimizer step.
\end{itemize}
Plotting $\Delta\mathcal{H}_2^{\text{true}}$ against $\Delta\mathcal{H}_2^{\text{pred}}$ over hundreds of steps yields a Pearson correlation above $0.95$ and a mean absolute error smaller than $0.02$ (for $\mathcal{H}_2$ typically in $[1.0, 4.0]$). This confirms that the first-order model accurately captures the direction and approximate magnitude of the entropy dynamics under the ICT sparse update regime.

\paragraph{Parameter sharing and AdamW.}
In neural networks, parameter updates affect many logits simultaneously. However, if $S$ contains the most unique tokens, their associated logits receive the dominant gradient signal, while other tokens experience only weak indirect changes. The L2 regularization inherent in AdamW further dampens extreme logit movements, preserving the near-independence assumed in our derivation. As a result, the unmodeled parameter-sharing effects remain negligible in practice.

\textbf{Conclusion}
Under the condition \(\pi(a) > e^{-1}\), both partial derivatives are negative:
\[
\operatorname{sgn}\left(\frac{\partial \mathcal{H}_1}{\partial \pi(a)}\right) = \operatorname{sgn}\left(\frac{\partial \mathcal{H}_2}{\partial \pi(a)}\right) = -1.
\]
Therefore, \(\mathcal{H}_1\) and \(\mathcal{H}_2\) exhibit homogeneity. This confirms that our method of updating unique tokens (which effectively modulates \(\mathcal{H}_2\)) also serves to control Shannon entropy (\(\mathcal{H}_1\)) in the same direction, preventing entropy explosion or collapse depending on the optimization regime.
\subsubsection{Detailed Algorithm Implementation}
\label{sec:detailed_algo}
In the main text, we present a concise version of the ICT framework (Algorithm~1). For reproducibility and to clarify implementation details, we provide a more elaborated pseudocode below. The detailed pseudocode incorporates several key refinements compared to the concise version in the main text: explicit handling of variable-length output sequences via the $\text{ValidSet}$ construction; precomputation of the group-average distribution $P_{\text{avg}}(t)$ for each time step $t$ to avoid redundant calculations; a clear definition of the percentile threshold using $100-k$ (where $k$ is the percentage of tokens to retain); and explicit freezing of the sampling policy by $\pi_{\text{old}} \leftarrow \theta$ at the beginning of each iteration.

\begin{center}
\noindent
\begin{minipage}[t]{0.65\textwidth}
    \vspace{0pt}
    \hrule height 0.8pt
    \vspace{3pt}
    \small \textbf{Algorithm 2} Detailed ICT framework
    \vspace{3pt}
    \hrule height 0.4pt
    \begin{algorithmic}[1]
        \small
        \STATE \textbf{Input:} $\mathcal{D}, G, \Theta, \alpha, k$ (retain top $k\%$, e.g., $k=10$)
        \STATE \textbf{Params:} Policy $\theta, \pi_{\text{ref}}$
        \FOR{step $s = 1, \dots, S$}
            \STATE $\pi_{\text{old}} \leftarrow \theta$ \COMMENT{Freeze the sampling policy}
            \STATE $q \sim \mathcal{D}$; $\{o_i\}_{i=1}^G \sim \pi_{\theta}(\cdot|q)$
            \STATE Compute rewards $r_i$ and advantages $A_i$
            \FOR{each time step $t$ (up to max output length)}
                \STATE $\text{ValidSet} = \{ j \mid \text{len}(o_j) \ge t \}$
                \STATE $P_{\text{avg}}(t) \leftarrow \frac{1}{|\text{ValidSet}|} \sum_{j \in \text{ValidSet}} \text{softmax}(L_{j,t})$
            \ENDFOR
            \FOR{$i \in \{1, \dots, G\}$}
                \STATE $\Psi_{i,t}(\theta) \leftarrow$ (PPO-Clip + KL) using $\pi_{\text{old}}$
                \IF{$s < \Theta$}
                    \STATE $M_{i,t} \leftarrow 1$ \COMMENT{Warm-up: update all tokens}
                \ELSE
                    \FOR{$t = 1$ to $T_i$}
                        \STATE $u_{i,t} \leftarrow D_{JS}(\text{softmax}(L_{i,t}) \parallel P_{\text{avg}}(t))$
                        \STATE $\text{Perc} \leftarrow \text{Percentile}(\{u_{i,\tau}\}_{\tau=1}^{T_i}, 100-k)$
                        \STATE $M_{i,t} \leftarrow \mathbb{I}[u_{i,t} \ge \text{Perc}]$
                    \ENDFOR
                \ENDIF
            \ENDFOR
            \STATE $\mathcal{L} \leftarrow - \frac{1}{G} \sum_{i=1}^G \frac{1}{\sum_t M_{i,t}} \sum_{t} M_{i,t} \Psi_{i,t}(\theta)$
            \STATE $\theta \leftarrow \theta - \alpha \nabla_\theta \mathcal{L}$
        \ENDFOR
    \end{algorithmic}
    \hrule height 0.9pt
\end{minipage}
\end{center}
\subsubsection{Implementation Details of the ICT Selector}
\label{app:ict_selector_details}

ICT selects tokens within each prompt group. For a prompt $q$, we sample a group of $G$ rollouts $\{o_i\}_{i=1}^{G}$ from the current policy. At each decoding position $t$, only rollouts whose lengths are at least $t$ are included in the valid set,
\begin{equation}
    \mathcal{V}_t = \{j \mid |o_j| \geq t\}.
\end{equation}
The group-average distribution at position $t$ is then computed as
\begin{equation}
    P_{\mathrm{avg}}(\cdot \mid t)
    = \frac{1}{|\mathcal{V}_t|}\sum_{j\in \mathcal{V}_t}
    \mathrm{softmax}(L_{j,t}),
\end{equation}
where $L_{j,t}$ denotes the full-vocabulary logits of rollout $j$ at position $t$. For each valid token position $(i,t)$, the uniqueness score is
\begin{equation}
    u_{i,t} = D_{\mathrm{JS}}\left(\mathrm{softmax}(L_{i,t}) \;\middle\|\; P_{\mathrm{avg}}(\cdot \mid t)\right).
\end{equation}
For each response $o_i$, ICT retains the top $k\%$ token positions according to $u_{i,t}$ and masks the remaining positions in the policy-gradient objective. Padding positions are excluded from both the computation of $P_{\mathrm{avg}}$ and the percentile threshold. Unless otherwise stated, we set $k=10$ after the warm-up phase.

In practice, the JS divergence is computed from batched logits that are already produced during rollout scoring. Therefore, the selector does not introduce an additional forward pass. The main extra cost is the computation and temporary storage of token-level probability vectors. This cost is amortized by batched tensor operations and is partially offset by the sparse backward pass, since gradients are retained only for selected token positions. We will report wall-clock time, peak GPU memory, and throughput in the final version to make this trade-off explicit.

\subsubsection{Clarification of the Sparse Objective}
\label{app:sparse_objective_clarification}

The sparse mask is applied to token-level policy-gradient terms after the GRPO advantage, clipping, and KL terms are computed. Thus, ICT does not alter the reward function, advantage estimator, reference policy, clipping rule, or KL regularization. It only changes which token positions contribute gradients to the policy update:
\begin{equation}
    \mathcal{L}_{\mathrm{ICT}}(\theta)
    = -\frac{1}{G}\sum_{i=1}^{G}
    \frac{1}{\sum_{t=1}^{T_i}M_{i,t}}
    \sum_{t=1}^{T_i}M_{i,t}\,\Psi_{i,t}(\theta),
\end{equation}
where $M_{i,t}\in\{0,1\}$ is the ICT mask and $\Psi_{i,t}(\theta)$ is the standard token-level GRPO objective containing the clipped probability ratio and KL penalty. When all $M_{i,t}=1$, the objective reduces to the dense GRPO update.

The theoretical analysis in the main text considers a positive logit update for an individual selected token in order to isolate the entropy effect of sparse token updates. In the actual GRPO objective, the sign and magnitude of the update are determined by the advantage, clipping, and KL penalty. Therefore, the derivation should be interpreted as a local diagnostic of how selected token positions can influence second-order entropy, rather than as a complete proof of the full optimizer dynamics.
\subsection{Ablation Analysis}
\label{Ablation Analysis}
\subsubsection{The best Unique Ratio}
\begin{figure}[ht]
  \vskip 0.2in
  \vspace{-0.7cm}
  \small
  \begin{center}
    \centerline{\includegraphics[width=\columnwidth]{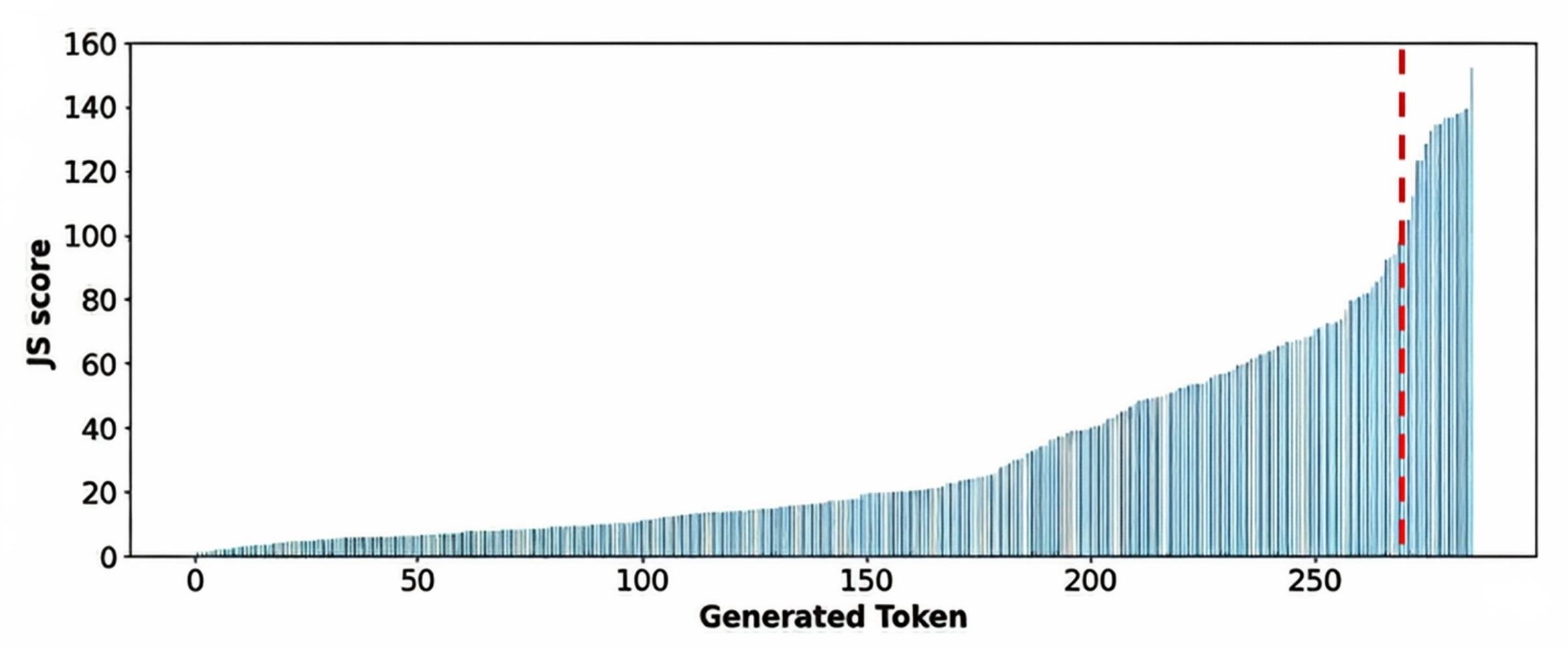}}
    \caption{
      Sorted Jensen-Shannon (JS) divergence scores with respect to the group-average distribution. The red dashed line marks the top 10\% threshold at the inflection point, where token uniqueness increases sharply. This empirically justifies our selection of unique tokens via the ICT distributional selector.
    }
    \label{fig:distribution}
  \end{center}
  \vspace{-0.3cm} 
\end{figure}
\begin{figure}[ht]
  \vskip 0.2in
  \vspace{-0.6cm}

  \begin{center}\centerline{\includegraphics[width=\columnwidth]{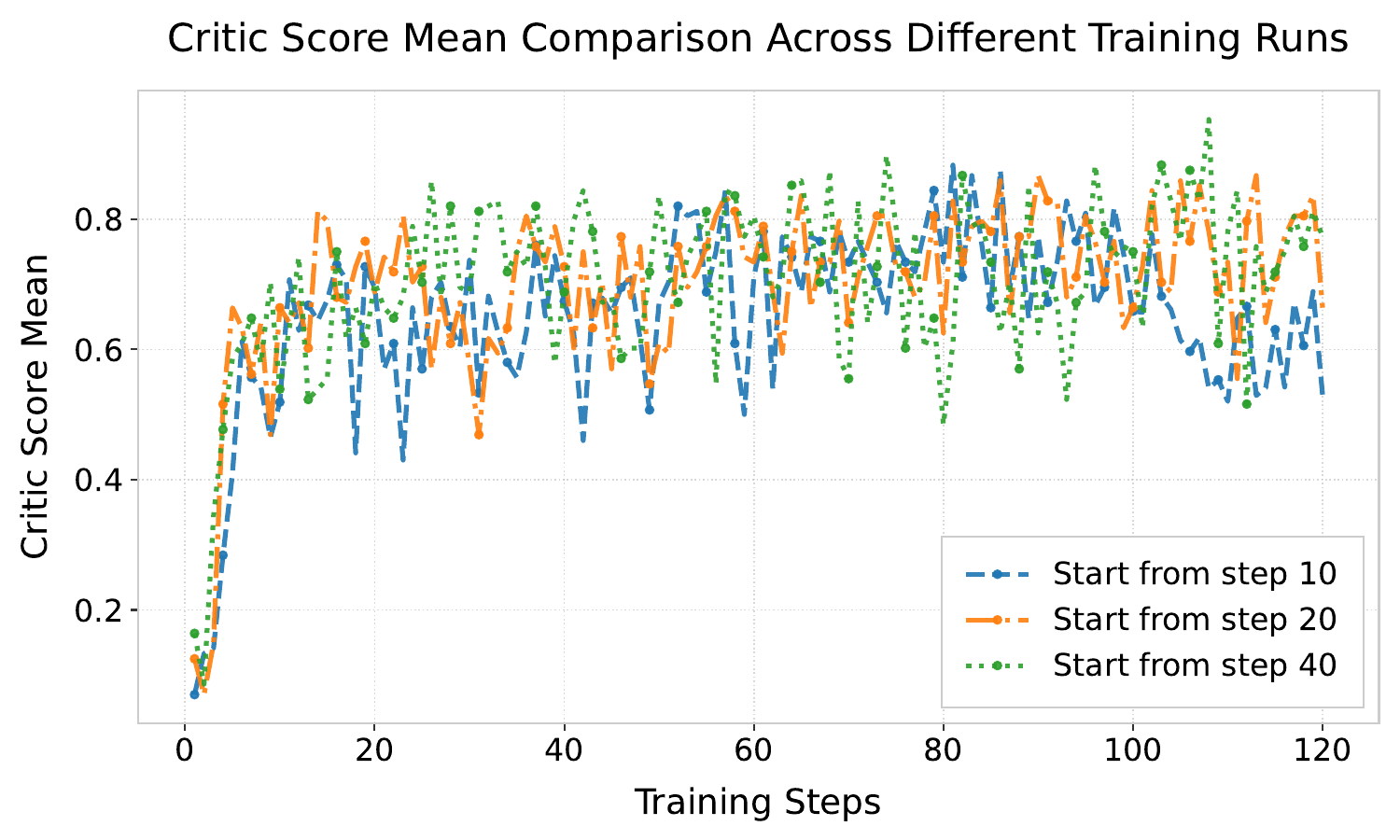}}
    \caption{
      Mean critic score trajectories for sparse updates activated after different warm-up lengths (step~10 in blue, step~20 in orange, and step~40 in green). Regardless of the warm-up duration, all configurations exhibit a rapid initial improvement followed by convergence to a comparable stable range (approximately 0.6--0.8). The substantial overlap in asymptotic behavior demonstrates the robustness of ICT to the timing of sparsity activation.
    }
    \label{fig:score} 
  \end{center}

\end{figure}
To validate our choice of sparsity ratio, we analyze the distribution of token uniqueness on the \texttt{Qwen2.5-1.5B} model using the GSM8K benchmark. Specifically, we compute the Jensen-Shannon (JS) divergence between the logits of each generated token and the group-average distribution. Sorting these tokens by their JS scores reveals a sharp inflection point, as illustrated in Figure~3. The distribution exhibits a pronounced elbow at approximately the top 10\% percentile. Beyond this threshold, the JS scores rise rapidly, indicating that this small subset of tokens exhibits substantial distributional deviation from the consensus. This finding confirms that token uniqueness is concentrated in a small fraction of critical branching points, thereby supporting our selection of the top 10\% as unique tokens for optimization via the ICT distributional selector.

\subsection{Selection Methods Comparison}
To compare different methods that can test the influence of distributional similarity, we change the top 10\% unique token selection from JS divergence to KL divergence, Wasserstein distance. KL divergence is inherently asymmetric ($D_{KL}(P \parallel Q) \neq D_{KL}(Q \parallel P)$), which makes it a biased estimator for token uniqueness. In the context of LLM logits, many critical tokens are characterized by strong deviations from the group average $P_{avg}$. 
Asymmetrically, KL divergence often focuses on tokens where the current probability $P_t$ is high relative to $P_{avg}$ (mode-seeking), while potentially ignoring tokens with large absolute logit deviations that are negative (i.e., tokens that the model strongly "rejects" compared to the consensus). 
Since JS divergence is symmetric and compares both distributions to a midpoint $M = \frac{1}{2}(P_t + P_{avg})$, it captures the \textit{total} distributional shift. This allows the ICT selector to identify any token with unique distributional properties—regardless of whether the deviation is positive or negative—as a critical logical branching point.

JS vs. Wasserstein Distance: Categorical vs. Geometric Alignment.
While Wasserstein distance (Earth Mover's Distance) is effective at capturing geometric shifts in continuous spaces, it is less suited for the discrete and categorical vocabulary space of LLMs. Wasserstein distance requires a predefined ground metric (distance between token indices), which is often semantically meaningless in a high-dimensional vocabulary. 
In contrast, ICT is fundamentally grounded in information theory and entropy dynamics (as discussed in Section 3.1). JS divergence is an entropy-based metric that naturally aligns with Shannon and Rényi entropy. It identifies uniqueness based on \textit{informational overlap} rather than \textit{transportation cost}. Since reasoning branching points are defined by their informational importance rather than geometric proximity in the embedding space, JS divergence provides a more stable and accurate signal for guiding RLVR exploration.

Consequently, JS divergence provides a more robust "uniqueness score" that prevents the gradients from being dominated by trivial tokens, thereby achieving higher Pass@1 and Pass@4 scores on the GSM8K benchmark.

To compare the different methods can be used to test the similarties of distribution' influence, we change the top 10\% unqiue token selection methods from JS divergence to KL divergence and Wasserstein distance. The GSM8K test results are shown in Table \ref{Different Unique token Selection Methods}. 
\begin{table}[h]
\caption{Different Unique token Selection Methods Comparatsion}
\centering
\label{Different Unique token Selection Methods}
\begin{tabular}{lcc}
\hline
Method   & GSM8K pass@1 & GSM8K pass@4 \\ \hline
ICT-Wass & 60.86        & 82.87        \\
ICT-KL   & 58.64        & 81.35        \\
ICT-JS   & 66.23        & 85.31        \\ \hline
\end{tabular}
\end{table}
Besides, we also train models by random choice method. we randomly choose 10\% tokens 
 to update, and test on GSM8K test dataset. The pass@1 is 63.65\%, pass@4 is 84.99\%, inferior to ICT method.

\subsubsection{Different Warm-step influence}
To investigate the sensitivity of the ICT framework to the warm-up duration, we vary the activation of sparse updates at steps 10, 20, and 40 (Figure~5). Baseline GRPO experiments indicate that the model typically reaches reward stability—corresponding to the emergence of strong reasoning capability—around step 20. Our results show that initiating sparse updates at step 10 leads to slightly slower convergence to the stable reward plateau, yet all three configurations eventually achieve comparable performance. This demonstrates the robustness of the ICT framework to the timing of sparsity activation, provided the model has sufficiently explored the initial policy landscape.
\subsubsection{Token Cloud Analysis}
\begin{figure}[ht]
\small
\vskip 0.05in
\begin{center}
\centerline{\includegraphics[width=0.8\columnwidth]{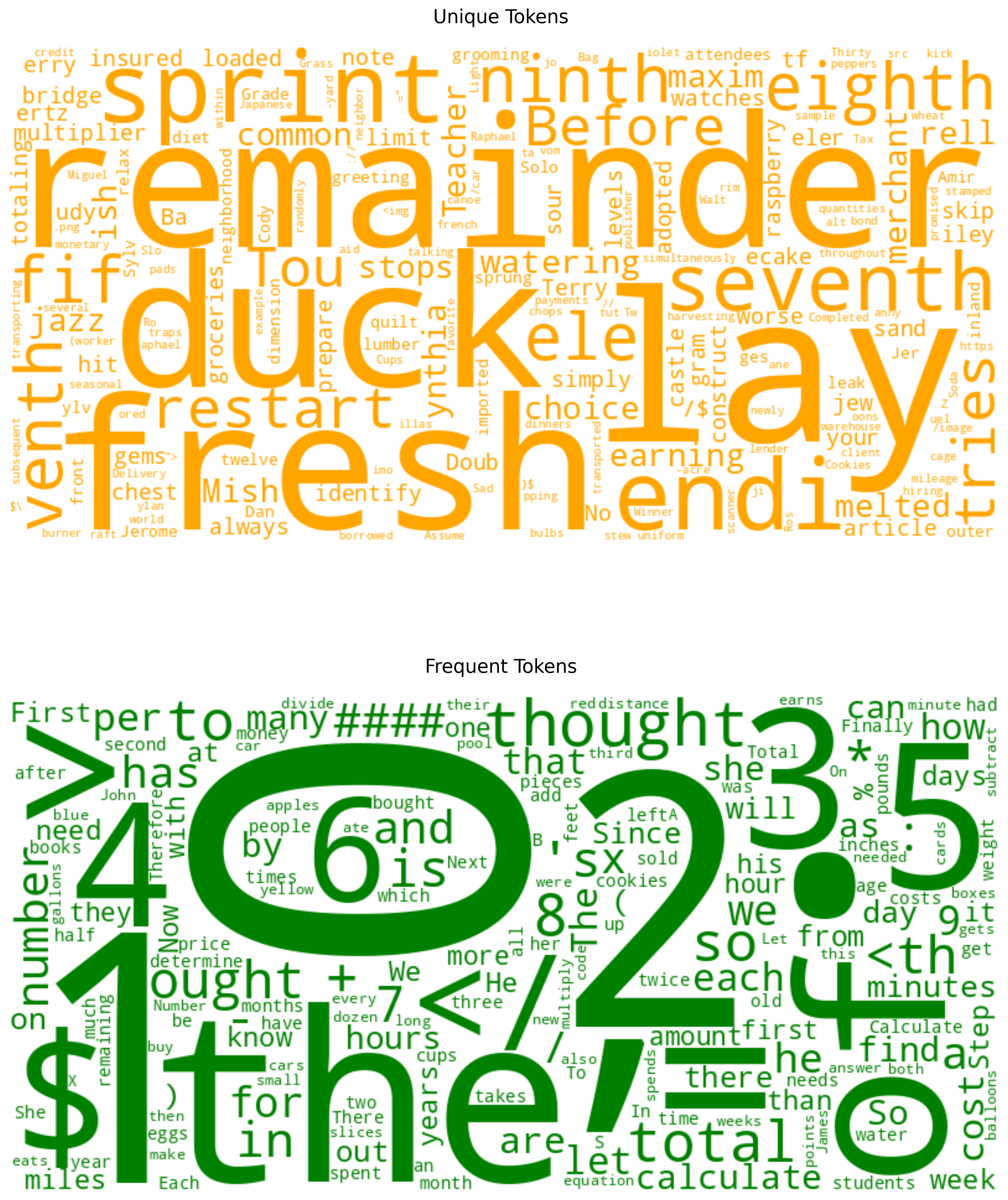}}
\caption{Word cloud visualization comparing unique tokens versus frequent tokens.}
\label{fig:token_cloud}
\end{center}
\end{figure}

\textbf{Qualitative Analysis of Token Roles.} To interpret the semantic distinction between selected and discarded tokens, we visualize the word clouds of unique (top 10\%) versus frequent tokens from GSM8K responses in Figure~\ref{fig:token_cloud}. Frequent tokens (bottom) are dominated by structural templates (e.g., ``\#\#\#\#'' for answer delineation) and common function words (e.g., ``the'', ``is''), which appear ubiquitously across samples. These tokens represent fixed reasoning scaffolding that requires minimal weight adjustment. In contrast, unique tokens (top) consist of problem-specific content words (e.g., ``remainder'', ``duck'', ``seventh''), indicating that our method successfully directs optimization toward exploring novel, context-dependent reasoning paths rather than reinforcing static formatting artifacts.

\subsection{Experiment Setting details}
\label{Experiment Setting details}
All ICT methods and the GRPO baseline share an identical optimization process, differing only in the criterion used to filter tokens for updates. We additionally include the 20-Entropy/80 method and STAPO as baselines. For the 20/80 method, the high-entropy threshold is set to 20\%, which is consistent with the original paper. For the STAPO method, the two key parameters are set to 0.002 for the probability threshold and 80\% for the entropy threshold. 

Training details.The 7B experiments are implemented within the veRL codebase and executed on a cluster of $4 \times$ NVIDIA H200 GPUs, while the 0.5B and 1.5B experiments are executed on a cluster of $4 \times$ NVIDIA V100 GPUs. For all tasks, each prompt is formatted with the instruction: ``Please reason step by step, and put your final answer within \boxed{}''. The maximum response length is set to 512 tokens for GSM8K and 2048 tokens for MATH, with a default learning rate of $1 \times 10^{-6}$,and the decoding temperature is 0.6. For the 7B experiments utilizing the Qwen2.5-7B base model on the H200 cluster, the actor learning rate is adjusted to $2 \times 10^{-6}$. These models are trained with a training and validation batch size of 16, resulting in a PPO mini-batch size of 16 and a micro-batch size of 4 across 2 PPO epochs. For each problem, we generate 8 rollouts with both the maximum prompt and response lengths limited to 2048 tokens. The advantage estimator is set to GRPO, complemented by a critic warmup of 5 steps and a math-based reward solver. To optimize memory usage on the H200 cluster (with GPU memory utilization set to 0.3), we utilize float16 precision, enable gradient checkpointing, and apply FSDP parameter offloading. Dynamic batch sizing is not applied. Other significant hyperparameters for the V100 configuration are provided in Appendix\ref{appendix:v100-hyperparams}.

\subsubsection{V100 Cluster Hyperparameters }
\label{appendix:v100-hyperparams}

The V100 cluster experiments use $4 \times$ NVIDIA V100 (32GB) GPUs with float16 precision.
Gradient checkpointing is enabled, and FSDP parameter offloading is applied to both the actor
and reference models. Dynamic batch sizing is disabled. The base model is Qwen2.5-0.5B, and
the reward solver is math-based (gsm8k). Each prompt follows the template ``Please reason step
by step, and put your final answer within \boxed{}''. The maximum prompt and response lengths
are 512 and 512 tokens for GSM8K, respectively. For each problem, 8 rollouts are generated with
a decoding temperature of 0.6 ($top$-$k = -1$, $top$-$p = 1.0$). The vLLM GPU memory utilization
is set to 0.3 with eager mode enabled. The actor and critic learning rates are
$1 \times 10^{-6}$ and $1 \times 10^{-5}$, respectively (constant schedule, no warmup), with
gradient clipping at 1.0. The advantage estimator is GRPO, with a PPO mini-batch size of 16,
micro-batch size of 4, and 2 PPO epochs per update. The PPO clip ratio is 0.2, and the critic
warmup consists of 5 steps. The KL penalty coefficient is 0.001 (fixed), and the entropy
coefficient is 0.001. The training and validation batch sizes are both 16.

\subsection{Statistical Reliability}
\label{app:statistical_reliability}

The current experiments are averaged over five random seeds. To make the reliability of the reported improvements clearer, we will add standard deviations or standard errors for the main results and ablations. We will also report paired significance tests across seeds where the same seed is used for ICT and the corresponding baseline. This is especially important for benchmarks such as AIME and GPQA, where the number of examples is limited and individual metric differences can have high variance.

\subsection{Scope of the Theoretical Analysis}
\label{app:theory_scope}

Our entropy analysis is intended to explain the local behavior of sparse token updates under simplifying assumptions. In particular, the derivation assumes a small update to a selected token logit and uses a first-order Taylor approximation. This abstraction omits several factors present in practical RLVR training, including multi-token parameter coupling, negative advantages, optimizer momentum, and so on. Consequently, the theory should be viewed as a mechanistic explanation for why distributionally sparse unique tokens can regulate exploration

\subsection{Broader Impact and Ethical Considerations}
\label{app:broader_impact}

ICT is a training method for improving the reasoning ability and training efficiency of large language models under verifiable rewards. The method does not introduce new data sources, user profiling, surveillance capabilities, or domain-specific harmful functionality. We therefore do not identify a direct ethical concern requiring special review.

However, stronger reasoning models can be dual-use. Improvements in mathematical and procedural reasoning may also increase the ability of models to assist with undesirable activities if deployed without safeguards. In addition, benchmark-driven reasoning optimization may amplify benchmark contamination concerns or encourage overfitting to narrow verifiable tasks. Responsible deployment should therefore combine ICT-trained models with standard safety evaluations, misuse monitoring, and domain-specific access controls when applied to sensitive settings. We will open-source the code and accept public review in the future.

\subsection{Limitations And Future Work}
\label{app:future_work}

Future work will investigate adaptive retained-token ratios, more efficient approximations to full-vocabulary JS divergence, stronger causal tests of selected token importance, and extensions to additional model families and domains. We also plan to study whether ICT can be combined with entropy regularization, process supervision, or verifier-guided search to further improve exploration while preserving stable training dynamics. And at present, we just use a fixed properties to select the unique tokens, we will use dynamical choice methods to select unqiue tokens based on the different distribution features in the future.


\bibliographystyle{plainnat}
\bibliography{example_paper}

\end{document}